\documentclass[10pt]{article} 
\usepackage{PRIMEarxiv}
\usepackage{natbib}

\usepackage{amsmath,amsfonts,bm}

\def\eqref#1{equation~\ref{#1}}

\def\1{\bm{1}}

\DeclareMathAlphabet{\mathsfit}{\encodingdefault}{\sfdefault}{m}{sl}
\SetMathAlphabet{\mathsfit}{bold}{\encodingdefault}{\sfdefault}{bx}{n}

\usepackage{wrapfig}
\usepackage{tikz}
\usepackage{subcaption}
\usepackage{booktabs}
\usepackage{multirow}

\usepackage{hyperref}
\usepackage{url}
\usepackage[noend]{algpseudocode}
\usepackage[linesnumbered,ruled,vlined]{algorithm2e}
\SetKwInput{Require}{Require}              
\SetKwInput{HyperParams}{Hyper-Parameters}        

\DeclareMathOperator{\conv}{conv}
\title{Memory Constrained Dynamic Subnetwork Update for Transfer Learning}

\author{Aël Quélennec
\qquad
Pavlo Mozharovskyi
\qquad
Van-Tam Nguyen
\qquad
Enzo Tartaglione
\\
LTCI, Télécom Paris, Institut Polytechnique de Paris\\
19 Place Marguerite Perey, 91120 Palaiseau, France\\
{\tt\small \{name.surname\}@telecom-paris.fr}}

\begin{document}

\maketitle

\begin{abstract}
On-device neural network training faces critical memory constraints that limit the adaptation of pre-trained models to downstream tasks. We present MeDyate, a theoretically-grounded framework for memory-constrained dynamic subnetwork adaptation. Our approach introduces two key innovations: LaRa (Layer Ranking), an improved layer importance metric that enables principled layer pre-selection, and a dynamic channel sampling strategy that exploits the temporal stability of channel importance distributions during fine-tuning. MeDyate dynamically resamples channels between epochs according to importance-weighted probabilities, ensuring comprehensive parameter space exploration while respecting strict memory budgets. Extensive evaluation across a large panel of tasks and architectures demonstrates that MeDyate achieves state-of-the-art performance under extreme memory constraints, consistently outperforming existing static and dynamic approaches while maintaining high computational efficiency. Our method represents a significant step towards enabling efficient on-device learning by demonstrating effective fine-tuning with memory budgets as low as a few hundred kB of RAM.
\end{abstract}

\section{Introduction}
\label{sec:introduction}

The exponential growth of deep neural networks has fundamentally transformed artificial intelligence, enabling breakthrough performances across diverse domains including Computer Vision~\cite{krizhevsky2012imagenet,simonyan2014very,minaee2021image}, Natural Language Processing~\cite{vaswani2017attention,tenney2019bert}, and Speech Recognition~\cite{deng2013new,nassif2019speech}. This remarkable progress stems from the continuous scaling of model complexity, with the number of parameters doubling every 8 to 17 months since the advent of the "Large Scale Era" marked by AlphaGo's release in 2015~\cite{sevilla2022compute}. While this trend further demonstrates the intrisic generalization potential of deep learning, it raises profound ecological and technical concerns. Training and exploitation of these architectures require very high energy consumption, and their deployment in real-world environments is impossible without extensive compression, leading to performance degradation and creating an increasingly critical tension between model capabilities and practical deployment constraints.\\
The proliferation of Internet of Things (IoT) devices and the growing demand for Edge AI applications have intensified the need for efficient on-device solutions. Traditional approaches to edge deployment follow a paradigm of offline training on powerful hardware, followed by model compression and deployment for inference-only applications on resource-constrained devices. This research domain encompasses five principal methodologies, namely quantization techniques for precision reduction, low-rank decomposition for parameter compression, architectural innovations for compact model design, knowledge distillation for teacher-student learning, and network sparsification (commonly referred to as pruning) for structural optimization~\cite{8253600,deng2020model}. However, this paradigm suffers from fundamental limitations as models trained on static datasets inevitably experience performance degradation when deployed in dynamic real-world environments due to \emph{data drift} phenomena~\cite{sahiner2023data}. The inability to adapt and learn continuously after deployment limits the practical viability of edge AI systems, particularly in applications requiring personalization, privacy preservation, and real-time adaptation to evolving data distributions~\cite{incel2023device}.\\
The primary obstacle preventing widespread adoption of on-device learning lies in the prohibitive computational and memory demands of backpropagation, as it requires storing intermediate activations and computing gradients, leading to memory requirements that can exceed device capabilities by several orders of magnitude. This has motivated exploration of alternative learning paradigms, including the Forward-Forward algorithm~\cite{hinton2022forward}, hyperdimensional computing~\cite{yang2023device}, and gradient-free optimization methods~\cite{pau2023suitability}. However, these approaches consistently underperform compared to traditional backpropagation-based techniques, creating a performance-efficiency trade-off that has yet to be satisfactorily resolved.\\
Recent advances in memory-constrained transfer learning have demonstrated that strategic subnetwork selection offers a promising path toward bridging this performance gap. Pioneering work by ~\cite{lin2022device} showed that fine-tuning within extreme memory budgets (256kB) is feasible through static subnetwork pre-selection. Improving on this concept, Quélennec~\emph{et~al.}~\cite{quélennec2025studytrainingdynamicsmemoryconstrained} propose to study the \textbf{Tra}ining \textbf{Dy}namics (TraDy) of network fine-tuning, thus yielding an approach that leverages the heavy-tailed behavior of stochastic gradients and the architectural consistency of layer importance across downstream tasks to dynamically resample channels between epochs, within pre-selected layers. Their method demonstrates that the synergistic combination of strategic layer pre-selection and dynamic channel selection enables state-of-the-art performance even under extreme memory constraints. Activation compression represents an orthogonal approach to memory-efficient training that targets the storage of intermediate representations rather than parameter selection, as demonstrated by Nguyen~\emph{et~al.}~\cite{nguyen2025beyond} who compress activations stored for backpropagation. Despite achieving considerable memory compression and computational speedup, this method does not address the fundamental question of which network components to train, defaulting to a heuristic that updates only the final layers.
Building upon Quélennec~\emph{et~al.} insights, this work presents a comprehensive extension of the TraDy strategy in the form of \textbf{Me}mory-constrained \textbf{Dy}namic subnetwork upd\textbf{ate} (MeDyate).
Our main contributions can be summarized as follows.
\begin{itemize}
    \item We propose an improved layer ranking by leveraging the the heavy-tailed behavior of stochastic gradient (Sec.~\ref{sec:alt_layer_ranking}).
    \item We demonstrate that, for a given downstream task, the channel topology remains stable during training (Sec.~\ref{sec:channel_behavior}).
    \item Given memory constraints, only a subset of relevant channels can be updated at a time. By dynamically adjusting the channels updated between epochs, performance improves compared to static approaches. We propose an adaptive algorithm that dynamically samples channels within key layers, assigning update probabilities based on their importance (Sec.~\ref{sec:stochasticity_algorithm}).
    \item We test MeDyate in typical transfer learning setups, under extreme memory constraints, observing that it can achieve state-of-the-art performance across multiple efficient architectures (Sec.~\ref{sec:main_results}). Our approach allows us to drastically reduce FLOPs, weight and activation memory during training while demonstrating superior performance compared to other similar strategies.
\end{itemize}
\section{Related Works}
\label{sec:sota}


\textbf{Activation Map Compression.} A critical observation for memory-efficient training is that activation maps (the outputs of each layer after non-linearity application) occupy significantly more memory space than parameters during backpropagation, as they are essential for computing weight derivatives~\cite{cai2020tinytl}. This insight has motivated a dedicated research direction focused on compressing activation maps using techniques adapted from weight compression literature, including quantization~\cite{10446393, pan2021mesa}, sparsification~\cite{pmlr-v119-kurtz20a, jiang2022back}, entropy encoding~\cite{georgiadis2019accelerating}, wavelet transform methods~\cite{NEURIPS2022_81f19c0e} and most notably, Nguyen~\emph{et al.}'s application of tensor decomposition strategies, allowing for drastic reduction of memory usage alongside acceleration of the training process~\cite{nguyen2025beyond}. While MeDyate is primarily designed as a strategy for surgically selecting which parameters to fine-tune, it naturally induces activation sparsity, thereby contributing to this research domain as well.\\

\textbf{Gradient Pruning.} The concept of gradient pruning describes the selection of a specific subnetwork to train during backpropagation, while the remainder of the network remains frozen. It differs from classical pruning by maintaining the complete network architecture intact during inference, while selectively modifying the backpropagation phase through criterion-based gradient computation. Gradient pruning can either be applied to reduce the memory footprint in constrained environments, or to accelerate training while preserving on-task performance~\cite{zhang2024gradientbasedparameterselectionefficient, bragagnolo2022update, li2023scotti, ye2020accelerating, mcdanel2022accelerating}. Within the transfer learning context, Lee~\emph{et~al.}~\cite{lee2022surgical} provide particularly relevant insights by framing gradient pruning as a regularization mechanism, demonstrating that network blocks exhibit task-dependent contributions to downstream performance. In their work, they observe that such contribution can either be constructive or destructive with respect to the task and is predicted by the ration of gradient norm to parameter norm.\\

\textbf{On-Device Learning.} Regarding our design of the MeDyate algorithm, three lines of work have largely influenced our approach, each presenting strategies for gradient pruning in memory-constrained environments applied to pre-trained architecture fine-tuning.\\
The foundational work by Lin~\emph{et~al.}~\cite{lin2022device} combine selective parameter updating, in the form of Sparse Update schemes (SU), alongside operator reordering and quantization-aware scaling to enable fine-tuning on extreme edge devices. While demonstrating that memory-efficient subnetworks can achieve acceptable downstream performance, SU suffers from significant practical limitations: determining adequate sparse configurations requires computationally expensive offline analysis of accuracy contributions followed by evolutionary search for each network-budget combination, and the resulting static selections are applied uniformly across all downstream tasks under the implicit assumption that optimal layer-channel configurations remain fixed throughout training.\\
Kwon~\emph{et~al.}\cite{kwon2024tinytrainresourceawaretaskadaptivesparse} addressed some of these limitations by enhancing adaptability across architectures, datasets, and memory budgets. They achieve this through Fisher information computation on downstream task activations to rank layers and channels, followed by reweighting based on parameter count and MAC operations. However, this approach introduces a fundamental contradiction as computing Fisher information for all network channels requires more memory than the gradient computation it seeks to optimize. Moreover, as for SU, the selected subnetwork remains static during fine-tuning.\\
Finally, Quélennec~\emph{et~al.}~\cite{quélennec2025studytrainingdynamicsmemoryconstrained} developed a theoretically-grounded framework for dynamic subnetwork selection, building on evidence-based analysis rather than empirical observations. Their work established that stochastic gradients exhibit heavy-tailed behavior during transfer learning and that layer importance remains architecturally consistent across downstream tasks, enabling a-priori layer selection. Within these pre-selected layers, they implemented random channel sampling that dynamically resamples between epochs, demonstrating consistent performance improvements over static selection approaches. While providing valuable theoretical insights into gradient distribution patterns and validating the benefits of adaptive selection strategies, we believe that their layer and channel selection strategies can be improved to yield increased on-task performance.
Building upon these foundational insights, our extended framework further analyzes the underlying dynamics of transfer learning in deep neural networks. We thus introducing MeDyate, an adaptive channel selection strategy that addresses the limitations of previous approaches while maintaining strict memory constraints for both parameters and activations.\\
\section{Method}
\label{sec:method}

\begin{figure}[t]
    \centering
    \includegraphics[width=0.95\linewidth]{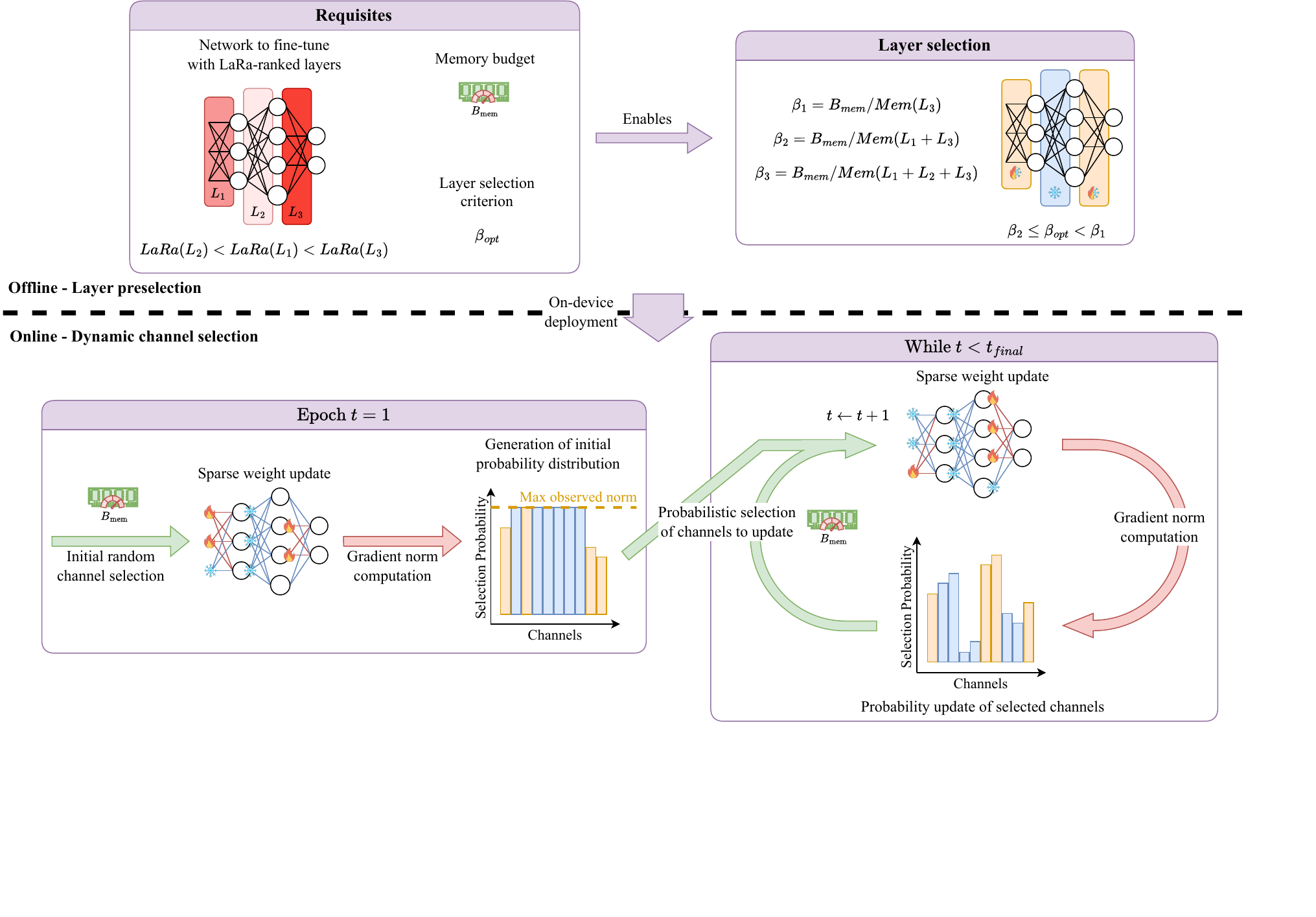}
    \caption{Overview of the MeDyate framework. The offline phase (top) shows layer pre-selection based on LaRa rankings and memory budget constraints, while the online phase (bottom) illustrates dynamic channel sampling with importance-weighted probabilities during training epochs.}
    \label{fig:MeDyate_recap}
    \vspace{-15pt}
\end{figure}

In this section, we present our comprehensive framework for parameter-efficient fine-tuning under extreme memory constraints. We begin by establishing the mathematical foundations and notation in Sec.~\ref{sec:prob_form}, providing the formal basis for our approach. In Sec.~\ref{sec:alt_layer_ranking}, we present an alternative methodology for layer ranking through the introduction of a more suitable metric, while Sec.~\ref{sec:channel_behavior} analyzes the stability of channel topology throughout training, demonstrating how channel importance distributions remain stable during the fine-tuning process. Building on these observations, Sec.~\ref{sec:stochasticity_algorithm} introduces our core dynamic channel sampling strategy, MeDyate, which enables efficient transfer learning within strict memory budgets through stochastic channel resampling between epochs. Finally, Sec.~\ref{sec:conv_and_compl_analysis} examines the properties of our algorithm in terms of both search space exploration and computational complexity.

\subsection{Notations}
\label{sec:prob_form}
Following the notation conventions established by Quelennec~\emph{et~al.} in their TraDy framework, we focus our mathematical formulation on 2D convolutional operations within CNNs, omitting bias parameters for clarity. It is however important to note that similar analysis naturally extends to fully-connected architectures. Our goal is to extend their methodological framework through theoretical analysis to improve upon their proposed layer and channel selection strategies. Thus, in a similar fashion to their work, we address the challenge of adapting pre-trained neural networks to downstream tasks within stringent memory budgets, operating without \emph{a priori} knowledge of the target domain. Our objective is to identify optimal subsets of the architecture for training, maximizing adaptation performance while maintaining both parameter and activation memory consumption within device constraints. The implicit assumption made here is that the memory constraint allows the device to perform the forward pass in its entirety, while the additional memory and latency requirements of backpropagation render full network training either impossible or impractical.\\
Let us then write the CNN as a composition of $n$ convolutional transformations:
\begin{equation}
    \mathcal{F}(\mathcal{X}) = \mathcal{C}_{\mathcal{W}_n} \circ \mathcal{C}_{\mathcal{W}_{n-1}} \circ \ldots \circ \mathcal{C}_{\mathcal{W}_1}(\mathcal{X}),
\end{equation}
where $\mathcal{X}$ represents the network input and $\mathcal{W}_i \in \mathbb{R}^{C' \times C \times D \times D}$ denotes the convolutional kernels for layer $i$, with $C$, $C'$ and $D$ indicating input, output channel and kernel dimensions respectively.\\
For any layer $i$, we define the input activation tensor $\mathcal{A}_{i} \in \mathbb{R}^{B \times C \times H \times W}$ and output tensor $\mathcal{A}_{i+1} \in \mathbb{R}^{B \times C' \times H' \times W'}$, where $B$ denotes batch size and $(H,W)$ represent spatial dimensions of the feature maps.\\
The gradient computation for layer weights follows standard backpropagation principles. The network loss $\mathcal{L}$ propagates backward from the output, generating activation gradients $\frac{\partial\mathcal{L}}{\partial\mathcal{A}_{i+1}}$ subsequently enabling weight derivative calculation:
\begin{equation}
    \frac{\partial\mathcal{L}}{\partial\mathcal{W}_i} = \conv\left(\mathcal{A}_i, \frac{\partial\mathcal{L}}{\partial\mathcal{A}_{i+1}}\right),
    \label{eq:weight_gradient}
\end{equation}
with the explicit tensor formulation:
\begin{equation}
    \left[\frac{\partial\mathcal{L}}{\partial\mathcal{W}_i}\right]_{c',c,k,l} = \sum_{b=1}^{B} \sum_{h'=1}^{H'} \sum_{w'=1}^{W'} \left[\mathcal{A}^{p}_{i}\right]_{b,c,h,w}\left[\frac{\partial\mathcal{L}}{\partial\mathcal{A}_{i+1}}\right]_{b,c',h',w'}, 
    \label{eq:explicit_gradient}
\end{equation}
where spatial indices follow $h = h' \cdot s + k \cdot d$ and $w = w' \cdot s + l \cdot d$ with stride $s$ and dilation $d$, and $\mathcal{A}_i^{p}$ represents the appropriately padded input tensor.\\
Consistent with the methodology established by Quélennec~\emph{et~al.} and Lin~\emph{et~al.} before them, we adopt input channels as the fundamental granularity for parameter selection. This choice is motivated by three key considerations. First, operating at finer granularities would produce unstructured sparse tensors during weight derivative computation, contradicting our memory efficiency objectives. Second, coarser granularities sacrifice selection precision by increasing the likelihood of scenarios where only a subset of parameters within a block requires updating while the majority remains irrelevant, leading to inefficient memory allocation. Third, and most importantly, input channel freezing generates natural activation sparsity alongside weight sparsity which is a property unique to this dimension. When an input channel is designated for freezing prior to a training epoch, we can proactively release the corresponding activation values from memory during forward propagation rather than storing them for potential use in backpropagation, thereby achieving additional memory savings. This advantage becomes particularly significant given that activation memory typically dominates the memory footprint during backpropagation compared to parameter storage, as demonstrated by Cai~\emph{et~al.}~\cite{cai2020tinytl}.\\
Throughout the remainder of this paper, we focus our analysis on two fundamental granularities for selective parameter updating: entire layers and individual input channels within those layers. These two levels enable efficient management of both weight and activation memory required for gradient computation. Moreover, as we will demonstrate, efficient layer selection (or exclusion, depending on perspective) proves crucial for reducing the search space of our proposed sampling methods.\\
Based on \eqref{eq:explicit_gradient}, we derive analytical expressions for the memory requirements and computational complexity associated with updating a single input channel $c$ within layer $i$. We define $\mathcal{C}^{\mathcal{W}_i}_c = C' \times D \times D$ as the weight memory cost and $\mathcal{C}^{\mathcal{A}_i}_c = H \times W$ as the activation memory cost for channel $c$. The total space complexity $(\Theta_{\text{space}})_c$ and time complexity $(\Theta_{\text{time}})_c$ for processing one input sample during backpropagation are as follows:
\begin{align}
    \left(\Theta_{\text{space}}\right)_c &= \mathcal{C}^{\mathcal{W}_i}_c + \mathcal{C}^{\mathcal{A}_i}_c,\label{eq:memory_cost}\\
    \left(\Theta_{\text{time}}\right)_c &= D^2 C' H' W'.
    \label{eq:computational_cost}
\end{align}

\subsection{Alternative Layer Ranking}
\label{sec:alt_layer_ranking}

In their work, Quélennec~\emph{et~al.} invoke two foundational studies to demonstrate that stochastic gradients are implicitly compressible. The first establishes that stochastic gradient noise exhibits heavy-tailed behavior during neural network training with stochastic gradient descent (SGD), as demonstrated by Şimşekli~\emph{et~al.}~\cite{simsekli2019heavy}. More precisely, the gradient noise $U_k(\mathcal{W})$ follows a symmetric $\alpha$-stable distribution:
\begin{equation}
\label{eq:heavy_tailed_sgd}
    U_k(\mathcal{W}) = \Delta\tilde{\mathcal{W}}_k - \Delta\mathcal{W} \sim S\alpha S(\sigma),
\end{equation}
where $\Delta\mathcal{W}$ represents the full-batch gradient, $\Delta\tilde{\mathcal{W}}_k$ the stochastic gradient computed from $k$ samples, and $\alpha \in (0,2]$ characterizes the tail heaviness (with decay proportional to $1/|x|^{\alpha+1}$). From \eqref{eq:heavy_tailed_sgd}, Quélennec~\emph{et~al.} observe that gradients naturally incorporate heavy-tailed noise components during stochastic optimization.\\
Complementing this theoretical foundation, Wan~\emph{et~al.} demonstrate that injecting heavy-tailed noise into weights during backpropagation renders them provably more compressible through pruning. The underlying mechanism is that heavy-tailed noise causes weight matrix columns to follow multivariate heavy-tailed distributions independently from each other. Consequently, the norm distribution becomes highly skewed: a small subset of columns exhibits disproportionately large norms while the majority remain relatively small. This concentration phenomenon means that the overall weight matrix norm is predominantly determined by a few dominant columns, creating an implicit sparse structure that aligns naturally with selective update requirements.\\
The natural conjunction of these two observations is that gradients, being composed of stochastic gradients and heavy-tailed noise, exhibit the same implicit compressibility properties described by Wan~\emph{et~al.}. Consequently, Quélennec~\emph{et~al.} propose to exploit input channel gradient norms to characterize update importance. More precisely, they reweight each channel's gradient norm by its associated memory cost as defined in~\eqref{eq:memory_cost}, producing a metric that combines both update relevance and memory efficiency. They denote this channel-level metric as RGN (Reweighted Gradient Norm):\\
\begin{equation}
    \text{RGN}_c = \frac{1}{\mathcal{C}^{\mathcal{W}_i}_c + \mathcal{C}^{\mathcal{A}_i}_c}\sqrt{\sum_{c',k,l}\left[\frac{\partial\mathcal{L}}{\partial\mathcal{W}_i}\right]^{2}_{c',c,k,l}} \text{.}
    \label{eq:reweighted_norm_channel}
\end{equation}
Quélennec~\emph{et~al.} then go on to hypothesize that layers can also be pruned from the training in a similar fashion to that of channels. In order to efficiently select which layers to freeze and which layers to update, they introduce a layer-related importance metric in the form of the sum of channels RGN:
\begin{equation}
    \left\| \frac{\partial\mathcal{L}}{\partial\mathcal{W}_i} \right\|_{\text{RGN}}= \frac{1}{\left(\Theta_{\text{space}}\right)_c}\sum_{c=1}^{C}{\left\|\left(\frac{\partial\mathcal{L}}{\partial\mathcal{W}_i}\right)_{c}\right\|_{2}} \text{.}
    \label{eq:layer_reweighting}
\end{equation}
However, we contend that this metric fails to capture the holistic behavior of entire layers. As the sum of individual channel gradient norms within a layer, the metric becomes dominated by channels with high values of high gradient norm, irrespective of how many channels exhibit comparatively low norms. Moreover, since the reweighting factor in~\eqref{eq:reweighted_norm_channel} corresponds to the memory cost of updating individual channels, and all channels within a layer share identical dimensions, this factor can be extracted from the summation as shown in~\eqref{eq:layer_reweighting}. Consequently, while this metric tends to favor layers containing channels that achieve favorable trade-offs between importance and update cost, it provides limited insight into overall layer behavior.\\
This limitation becomes particularly problematic in our framework, where layer selection and exclusion represent complementary aspects of the same decision. When a layer is excluded, it remains frozen throughout the entire training duration; conversely, when selected, we can reasonably assume complete layer coverage over time through the stochastic sampling approaches introduced by Quélennec~\emph{et~al.} and extended in this work. Given this binary layer treatment, the metric in~\eqref{eq:layer_reweighting} essentially characterizes individual channel behavior within layers rather than providing a comprehensive assessment of layer-level significance.\\
To address this limitation, we propose a layer-level metric that incorporates both the actual norm of the layer gradient and its complete memory cost. We denote this metric as LaRa (\textbf{La}yer \textbf{Ra}nking) and define it as:
\begin{equation}
\label{eq:layer_importance}
    \text{LaRa}_i = \frac{\left\|\frac{\partial\mathcal{L}}{\partial\mathcal{W}_i}\right\|_2}{\sum_{c=1}^{C}\left(\mathcal{C}^{\mathcal{W}_i}_c + \mathcal{C}^{\mathcal{A}_i}_c\right)},
\end{equation}
where $\left\|\frac{\partial\mathcal{L}}{\partial\mathcal{W}_i}\right\|_2$ represents the Euclidean norm of the entire layer gradient tensor. Beyond providing improved layer characterization, this metric preserves the property of layer ranking stability during fine-tuning across downstream tasks, as the theoretical inequalities established in prior work are based on Euclidean norms. In the experimental section (Sec.~\ref{sec:prel_exp}), we empirically validate that this new metric enables similar or superior performance with fewer layers while respecting memory budget constraints. As demonstrated in Sec.~\ref{sec:stochasticity_algorithm}, this reduction in the number of layers requiring updates proves crucial for the design of our sampling algorithm.\\

\subsection{Stability of Channel Importance Throughout Training}
\label{sec:channel_behavior}

A fundamental insight from the TraDy framework is that channel importance distributions (also termed "channel topology" by Quélennec~\emph{et~al.}) exhibit task-dependent variations that prevent \emph{a priori} channel selection across different downstream tasks. This task-specific property implies that static pre-computed subnetwork selection will underperform compared to task-adaptive approaches. However, in memory-constrained environments that prevent full gradient computation, reliable estimation of channel importance remains a complex challenge.\\
As established by Quélennec~\emph{et~al.}, the distribution of channel gradient norms varies significantly between downstream tasks. This observation stems from the fundamental composition of weight derivatives as expressed in~\eqref{eq:explicit_gradient}, where both activation maps and activation derivatives are inherently shaped by task-specific data characteristics. The activation maps $\mathcal{A}_i$ capture features extracted at each network layer and reflect how the network processes input data, while activation derivatives $\frac{\partial\mathcal{L}}{\partial\mathcal{A}_{i+1}}$ encode task-specific loss landscape information that varies according to the downstream objective.\\
We observe, however, that channel topology exhibits remarkable stability throughout training within a given downstream task. More precisely, this temporal consistency unfolds through two subsequent phases:
\begin{itemize}
    \item \textbf{Rapid Stabilization:} Channel importance distributions evolve quickly during initial training epochs but then stabilize for the remainder of the fine-tuning process. This rapid convergence suggests that the relative importance of channels becomes established early in adaptation and remains consistent as the network refines its task-specific representations.
    \item \textbf{Distributional Consistency:} Once stabilized, the channel topology maintains its structure throughout training, with channels preserving their relative importance rankings even as absolute gradient magnitudes may fluctuate during optimization (typically, average gradient norms decrease throughout training as a result of loss minimization).
\end{itemize}
This behavior can be explained by the fundamental nature of fine-tuning pre-trained architectures. During fine-tuning, the most significant weight modifications occur in the initial stages as the network adapts to the new loss landscape. Since fine-tuning typically assumes that downstream task features are sufficiently similar to those of the pre-training task, the network requires only targeted adaptations rather than complete relearning. In the first few epochs (or steps, depending on task similarity), loss and gradient magnitudes are relatively high as the network adjusts to new features, resulting in substantial weight changes that cause channel importance fluctuations.\\
Subsequently, as the network converges toward a local minimum, the rate of loss reduction decreases and absolute gradient values become smaller. Consequently, we observe overall weight stabilization with only minor adjustments. This stability propagates through the computational graph: as weights $\mathcal{W}_i$ stabilize and the input dataset remains fixed, the activations $\mathcal{A}_i$ also achieve stability. This stability then cascades to activation derivatives and ultimately to weight derivatives, which represent combinations of these stabilized components as expressed in~\eqref{eq:weight_gradient}.\\
This temporal stability has profound implications for memory-constrained channel selection. Since channel importance distributions remain consistent within tasks after initial stabilization, we can reliably estimate the overall channel topology through sampling approaches, even when memory constraints prevent computing gradients for all channels simultaneously.\\
Furthermore, in the TraDy framework Quélennec~\emph{et~al.} showcased that dynamic channel selection strategies consistently outperform their static counterparts under memory constraints. This superiority of dynamic approaches naturally aligns with our temporal stability observations: if channel importance distributions remain stable over time, then dynamic sampling can effectively explore this distribution while respecting memory limitations, ultimately approximating the performance approaches that would have full gradient knowledge.\\
The combination of temporal stability and dynamic sampling superiority forms the theoretical foundation for our dynamic channel selection strategy. By leveraging the stable channel topology within tasks, we can design sampling algorithms that efficiently explore channel importance space while maintaining strict memory constraints, confident that our estimates will remain representative throughout the training process.\\

\subsection{Dynamic Chanel Sampling}
\label{sec:stochasticity_algorithm}

The insights developed in the previous subsections directly inform the design of our dynamic channel sampling algorithm. The demonstrated stability of channel importance distributions within tasks enables us to dynamically sample channels for updating across epochs, yielding a faithful estimation of the underlying channel topology over time. Simultaneously, our alternative layer ranking methodology allows us to significantly reduce the search space for sampling by excluding inefficient layers \emph{a priori}, thereby accelerating convergence as the memory budget is concentrated within layers known to be more beneficial for adaptation.\\
Building upon these foundations, we introduce several algorithmic innovations. First, while Quélennec~\emph{et~al.} employed reweighted gradient norms as channel importance metrics, we contend that our LaRa-based layer exclusion enables a shift to raw gradient norms as channel importance indicators within the selected layer subset. Since LaRa already incorporates memory cost reweighting at the layer level, inefficient high-memory layers are excluded, allowing raw channel norms to serve as more direct measures of importance within the remaining efficient layers. Second, rather than deterministically selecting channels based on importance rankings, we propose a probabilistic selection strategy where each channel receives an update probability proportional to its importance. This stochastic approach ensures that all channels within selected layers receive updates at some point during training, while channels with higher importance naturally receive more frequent attention, leading to more robust and comprehensive network adaptation.\\
While our LaRa metric provides layer rankings, it does not inherently specify a stopping criterion for layer selection. Quélennec~\emph{et~al.} employed a fixed number of layers regardless of memory budget constraints, determined by ranking layers according to their RGN values and applying a threshold that captures $97\%$ of the cumulative gradient norm on a given downstream task. However, we argue that our more sophisticated channel selection strategy necessitates careful adaptation of the layer count $K$ based on the available memory budget. This requirement stems from the need to balance exploration and exploitation: we must adequately capture channel topology within few epochs while avoiding excessive search space reduction that would permanently exclude relevant channels from consideration.\\
To address this challenge, we introduce a hyperparameter $\alpha_K$ that represents the ratio between the total memory budget $B_{\text{mem}}$ and the memory footprint $M_K$ of the $K$ selected layers constituting our search space:
\begin{equation}
    \alpha_K = \frac{B_{\text{mem}}}{M_K}.
    \label{eq:layer_hyperparam}
\end{equation}
This formulation enables principled adjustment of the layer search space based on memory constraints, ensuring that our sampling strategy operates within an appropriately sized parameter space. In Sec.~\ref{sec:prel_exp}, we detail the systematic exploration conducted to determine the optimal value $\alpha_{\text{opt}}$. Given this optimal proportion and a specific memory budget, we can then adaptively select the $K_{\text{opt}}$ layers for our search space according to the ranking established by our LaRa metric.\\
We present MeDyate, our dynamic subnetwork update pipeline for memory-constrained transfer learning, as detailed in Algorithm~\ref{alg:adaptive_selection}. Prior to training, the top $K$ layers constituting the search space are selected such that $\alpha_K$, as defined in~\eqref{eq:layer_hyperparam}, represents the largest value not exceeding $\alpha_{\text{opt}}$ (line~\ref{line:topk_layers_select}).\\
The training process then proceeds as follows. Given a pre-trained backbone and training dataset, channels are initially sampled uniformly at random within the predefined layer set $L_K$, subject to the memory budget constraint (line~\ref{line:random_sampling}). In the second epoch, we compute the gradient norms of previously selected channels (line~\ref{line:grad_compute}) and update the sampling distribution by assigning corresponding norm values to sampled channels while setting unselected channels to the maximum observed norm value (line~\ref{line:set_maxnorm_proba}). As established in Sec.~\ref{sec:stochasticity_algorithm}, gradient magnitudes typically decrease throughout fine-tuning. Consequently, by assigning the maximum observed norm to unseen channels, we effectively grant them maximum selection probability, thereby promoting early exploration of the channel space.\\
Subsequently, channels are resampled according to a probability distribution proportional to the constructed gradient norm vector, again respecting the memory budget (line~\ref{line:resample}), and the selected channels are updated (line~\ref{line:update_weight}). From the third epoch onward, the maximum norm assignment step (line~\ref{line:set_maxnorm_proba}) is omitted, and we iteratively resample and update channels according to the probability distribution derived from the established norm vector. Upon completion of training, we evaluate model performance on the test dataset (line~\ref{line:test}).\\
Fig.~\ref{fig:MeDyate_recap} illustrates the operational flow of our MeDyate algorithm across its two distinct phases. In the offline phase, we assume the LaRa metric has been pre-computed through a preliminary fine-tuning run on available downstream task data. This pre-computation is justified by the stability of layer rankings across tasks, requiring only a single mock training session of a few epochs on any relevant downstream task to establish the layer importance hierarchy.\\
The online phase proceeds through three key stages. First, channels are initially selected uniformly at random within the pre-selected layer subset to respect memory budget constraints. Second, the probability distribution is instantiated by assigning gradient norm-proportional probabilities to previously selected channels while setting unobserved channels to the maximum observed probability to encourage exploration. Finally, the algorithm enters an iterative loop alternating between probabilistic channel selection based on the current distribution and probability distribution updates using newly computed gradient norms from the selected channels.\\

\begin{algorithm}[t]
\caption{MeDyate}
\label{alg:adaptive_selection}
\SetAlgoNoLine
\DontPrintSemicolon
  \Require{Pre-trained backbone weights $\mathcal{W}$, LaRa ranked layers $L$, initial empty channel norm vector $N$, number of epochs $n$, Train data $D_{\text{train}}$, Test data $D_{\text{test}}$, memory budget $B_{\text{mem}}$.}
  \HyperParams{Top $K$ layers selection criterion $\alpha_{\text{opt}}$.}
    \SetAlgoVlined
    \textbf{Initialization:} Select the top $K$ layers $L_K$ such that $\alpha_{K-1} > \alpha_{\text{opt}}$ and $\alpha_{K} \leq \alpha_{\text{opt}}$.\; \label{line:topk_layers_select}
    \For{$\text{epoch} = 1, ..., n$} 
    {   
        \uIf{$\text{epoch} = 1$}{
            Randomly sample channels $C_t$ within $L_K$ along uniform probability distribution until the memory budget $B_{\text{mem}}$ is met.\; \label{line:random_sampling}
          }
          \uElseIf{$epoch > 1$}{
            Compute gradient norm $\left\|\left(\frac{\partial\mathcal{L}}{\partial\mathcal{W}_i}\right)_{c}\right\|_{2}$ of channels selected at previous epoch.\;
            Update the corresponding norm values in $N$.\; \label{line:grad_compute}
            \uIf{$\text{epoch} = 2$}{
                Set the norm value for non-selected channels $\bar{C_t}$ to be the maximum norm value observed, $\max_c\left[\left\|\left(\frac{\partial\mathcal{L}}{\partial\mathcal{W}_i}\right)_{c}\right\|_{2}\right]$.\; \label{line:set_maxnorm_proba}
            }
            Resample channels $C_t$ along probability distribution proportional to $N$ until the memory budget $B_{\text{mem}}$ is met.\; \label{line:resample}
          }
        Update the weights of the selected channels $C_t$ using $D_{\text{train}}$.\; \label{line:update_weight}
    }
    Evaluate the fine-tuned backbone using $D_{\text{test}}$\; \label{line:test}
    \vspace{-1mm}
\end{algorithm}

\subsection{Convergence and Complexity Analysis}
\label{sec:conv_and_compl_analysis}
\begin{wrapfigure}{t}{0.33\textwidth}
    \centering
    \includegraphics[width=\linewidth,trim={0 0 0 0},clip]{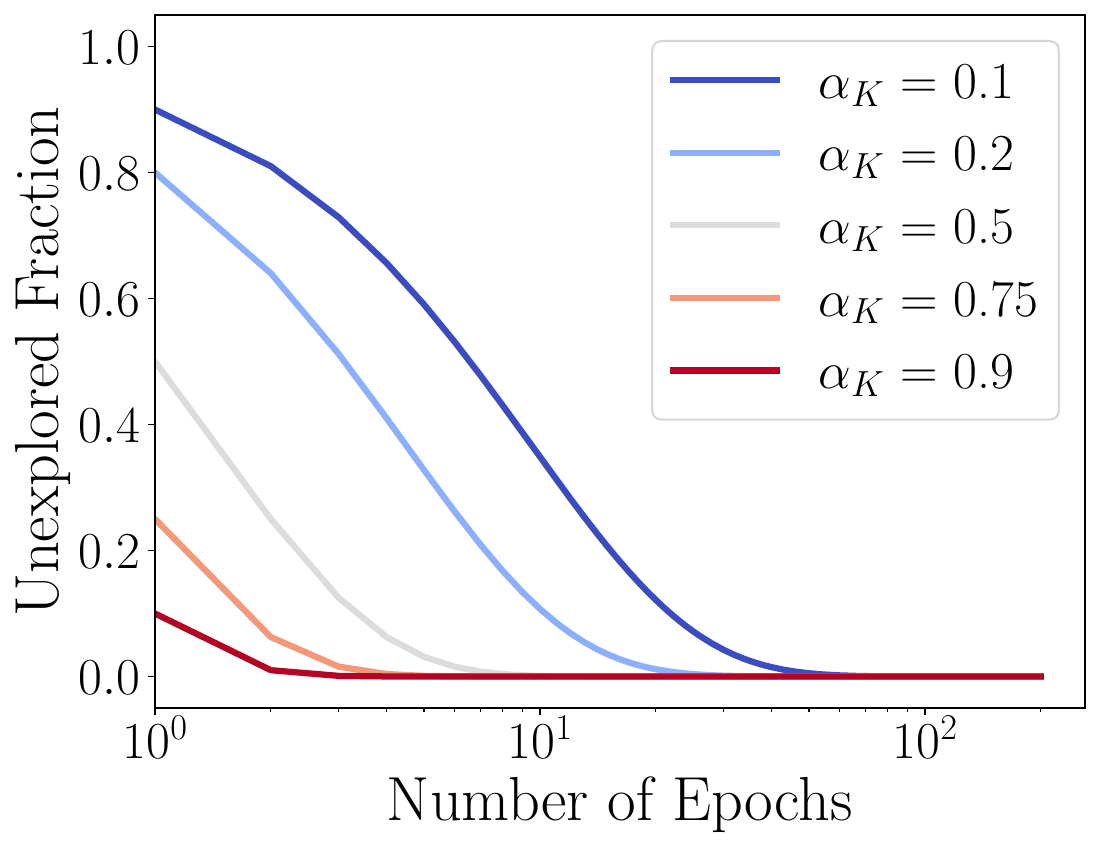}
    \caption{Evolution of $u_e$ as a function of epochs for different $\alpha$ values.}
    \label{fig:unexplored_alpha_epochs}
\end{wrapfigure}
Here we analyze the number of epochs required for our MeDyate algorithm to explore the search space and provide an analysis of the computational overhead of our method.

\textbf{MeDyate Convergence.} Due to our assignment of maximum probability to unseen channels, we can assert that our algorithm exhibits superior search space exploration compared to a fully random approach (i.e., where every channel receives equal selection probability). We therefore propose to analyze this random approach as a worst-case scenario bound for MeDyate convergence regarding exploration potential.\\
At each epoch, channels are selected according to a uniform distribution such that the memory budget $B_{\text{mem}}$ is satisfied. By design, this memory budget represents a proportion $\alpha_K$ of the total memory available within the search space. If we denote $u_e$ as the proportion of unexplored search space at epoch $e$, we naturally obtain $u_e = (1 - \alpha_K)^e$. This expression clearly highlights the interdependence between the number of selected layers and the memory budget as key factors determining our algorithm's capacity to explore the available search space.\\
Fig.~\ref{fig:unexplored_alpha_epochs} illustrates the evolution of the unexplored fraction as a function of epochs for different values of $\alpha$. In practice, we have $\alpha_K \simeq \alpha_{\text{opt}}$, and our hyperparameter search detailed in Sec.~\ref{sec:prel_exp} yields $\alpha_{\text{opt}} = 0.2$. Under random search conditions, this implies that $80\%$ of channels are observed within $8$ epochs, $90\%$ within $11$ epochs, and $95\%$ within $14$ epochs. As previously stated, these values represent upper bounds for exploration time, since we expect MeDyate to surpass random search performance. These rapid convergence rates are crucial, as they demonstrate that within just a few epochs, MeDyate can construct a comprehensive gradient representation, enabling it to achieve performance comparable to methods with full gradient knowledge.
\begin{figure*}[t]
    \begin{subfigure}{0.33\textwidth}
        \includegraphics[width=0.9\linewidth]{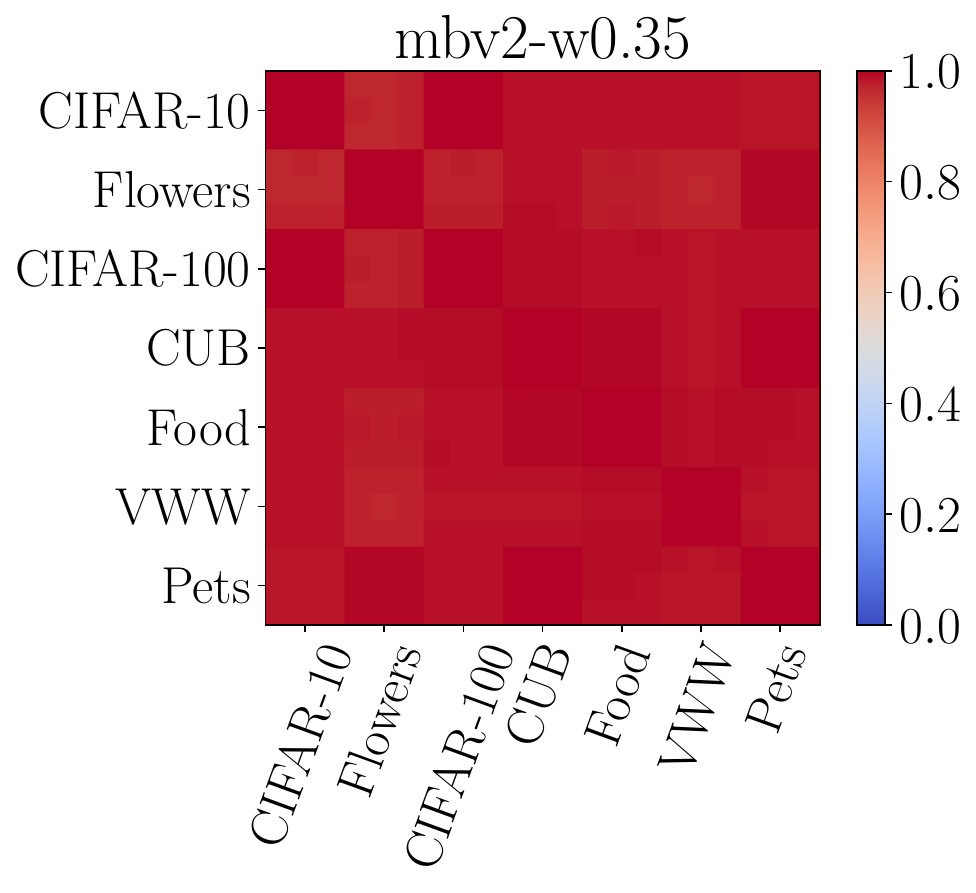}
        \label{fig:subim_mbv2_reord_sim}
    \end{subfigure}
    \begin{subfigure}{0.33\textwidth}
        \includegraphics[width=0.9\linewidth]{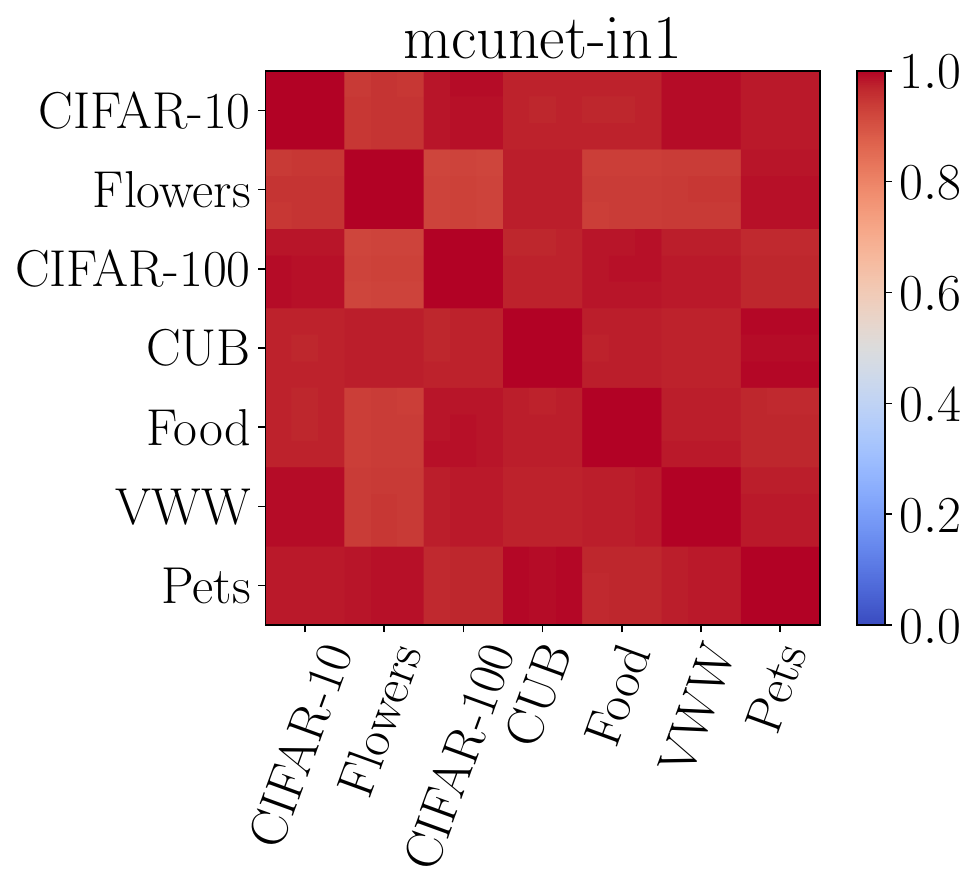}
        \label{fig:subim_mcunet_reord_sim}
    \end{subfigure}
    \begin{subfigure}{0.33\textwidth}
        \includegraphics[width=0.9\linewidth]{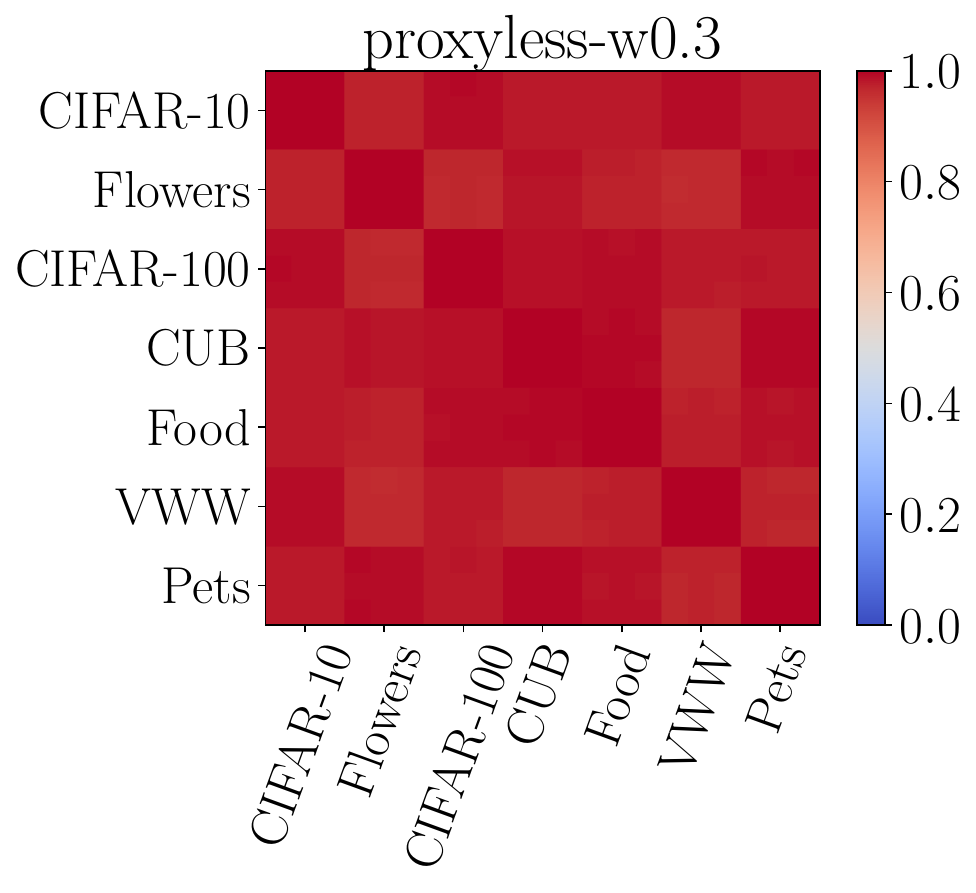}
        \label{fig:subim_proxyless_reord_sim}
    \end{subfigure}
    
    \caption{Spearman correlation of layer cumulated LaRa metric across seeds and datasets.}
    \label{fig:layer_sim}
\end{figure*}

\textbf{MeDyate Computational Overhead.} Compared to Lin~\emph{et~al.}'s SU approach, which adds no on-device computational overhead since update schemes are pre-computed, or Quélennec~\emph{et~al.}'s TraDy method, which performs random channel selection within predefined layer subsets, our MeDyate algorithm introduces more substantial computational overhead due to the necessity of computing gradient norms between epochs. We analyze this added complexity below.\\
For a given epoch $e$, let $\zeta_e$ denote the set of channels selected for update and $|\zeta_e|$ its cardinality. Assuming uniform channel dimensions for simplicity, the computational complexity $(\Theta_{\text{time}})^{RN}_e$ of computing norm metrics for selected channels between epochs is:
\begin{equation}
     (\Theta_{\text{time}})^{RN}_e = |\zeta_e|C'D^2.
     \label{eq:norm_complexity}
\end{equation}
The complexity of sampling one channel within a probability distribution of length $N$ is $\mathcal{O}(\log(N))$, making the complexity of sampling $|\zeta_{e+1}|$ channels $\mathcal{O}(|\zeta_{e+1}|\log(N))$. Since $|\zeta_{e+1}| \simeq |\zeta_e|$ in practice, the total computational complexity of MeDyate for epoch $e$, denoted $(\mathcal{O}_{\text{time}})^{\text{Med}}_e$, becomes:
\begin{equation}
	(\mathcal{O}_{\text{time}})^{\text{Med}}_e = |\zeta_e|(C'D^2 + \log(N)).
    \label{eq:med_complexity}
\end{equation}
To contextualize this overhead, we compare it to the computational cost of computing weight derivatives for selected channels during backpropagation. Let $M$ denote the number of steps per epoch. The computational complexity during epoch $e$, $(\Theta_{\text{time}})^{\text{grad}}_e$, is:
\begin{equation}
    (\Theta_{\text{time}})^{\text{grad}}_e = M|\zeta_e|D^2C'H'W'.
    \label{eq:grad_complexity}
\end{equation}
Furthermore, forward propagation and loss backpropagation exhibit similar computational complexity and involve the entire network rather than just the updated channel subset. Given the typical orders of magnitude of these variables, MeDyate's computational overhead remains negligible compared to the overall backpropagation cost. We anticipate a modest increase in inter-epoch latency, creating a trade-off between improved performance and computational overhead.

\section{Experiments}
\label{sec:experiments}

\subsection{Preamble}
\label{sec:preamble}

\textbf{Experimental Setup.} Our evaluation leverages three efficient architectures with ImageNet~\cite{deng2009imagenet} pre-training: MobileNetV2~\cite{sandler2018mobilenetv2}, ProxylessNAS~\cite{cai2018proxylessnas}, and MCUNet~\cite{lin2020mcunet}, adopting the same models used by Lin~\emph{et~al.}~\cite{lin2022device} and Quélennec~\emph{et~al.} for consistency. All experiments are conducted on Nvidia Tesla V100 SXM2 hardware using PyTorch 2.0.0 implementation in Python. Across all experimental configurations, we maintain classifier layer training regardless of the specific parameter selection strategy employed. Additional evaluation on transformer-based architectures is presented in the appendix.

\textbf{Evaluation Datasets.} We assess our approach across seven downstream tasks: CIFAR-10~\cite{krizhevsky2009learning}, CIFAR-100~\cite{krizhevsky2009learning}, CUB~\cite{cub}, Flowers~\cite{nilsback2008automated}, Food~\cite{bossard2014food}, Pets~\cite{parkhi2012cats}, and VWW~\cite{chowdhery2019visual}.\footnote{Pets: https://creativecommons.org/licenses/by-sa/4.0/, CC BY-SA 4.0 license; ImageNet: https://image-net.org/download.php the ImageNet license; others are not listed.} We reproduce Quélennec~\emph{et~al.} training protocol, employing cosine annealing schedules with 5-epoch warm-up periods~\cite{goyal2017accurate}, spanning 200 epochs for smaller datasets (CUB, Pets, Flowers), 100 epochs for CIFAR-100, and 50 epochs for larger datasets (CIFAR-100, Food, VWW). Learning rates decay from $0.125$ to $0$ without weight decay or dropout regularization. Statistical robustness is ensured through triple-seed evaluation, with reported metrics representing means and standard deviations across runs.

\textbf{Memory Budget Configuration.} To ensure fair comparative evaluation, we implement identical memory constraints to those established in Lin~\emph{et~al.}'s SU framework and reproduced in Quélennec~\emph{et~al.}'s TraDy framework. Our experimental design incorporates five graduated memory limitation tiers for each evaluated architecture, where each budget $B_{\text{mem}}$ defines the total permissible memory allocation encompassing both parameter updates and activation storage requirements during training. This standardized constraint methodology enables direct performance comparisons while ensuring that observed improvements stem from algorithmic advances rather than relaxed memory limitations, thereby providing rigorous validation of our theoretical contributions under equivalent resource restrictions.\\

\begin{figure*}[t]
    \begin{subfigure}{0.33\textwidth}
        \includegraphics[width=0.9\linewidth]{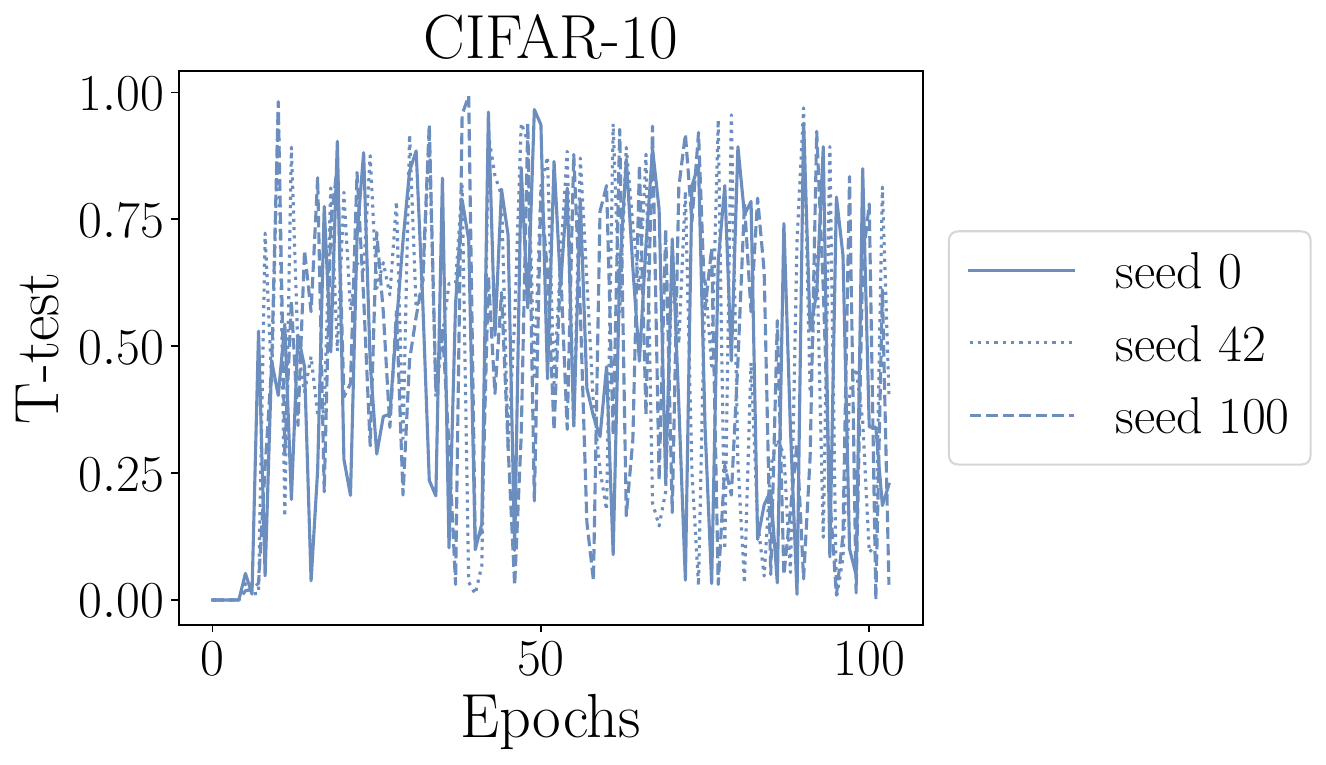}
        \label{fig:subimc10_ttest_mbv2}
    \end{subfigure}
    \begin{subfigure}{0.33\textwidth}
        \includegraphics[width=0.9\linewidth]{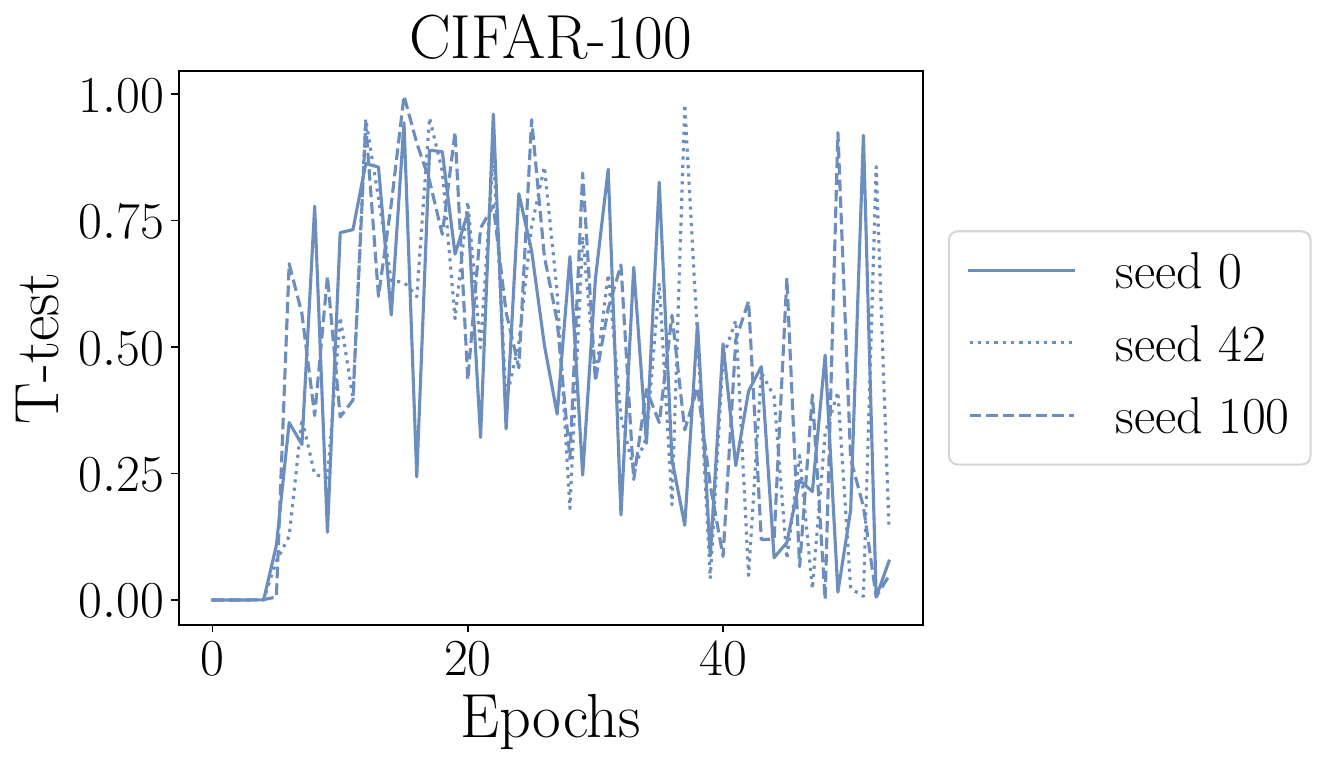}
        \label{fig:subimc100_ttest_mcunet}
    \end{subfigure}
    \begin{subfigure}{0.33\textwidth}
        \includegraphics[width=0.9\linewidth]{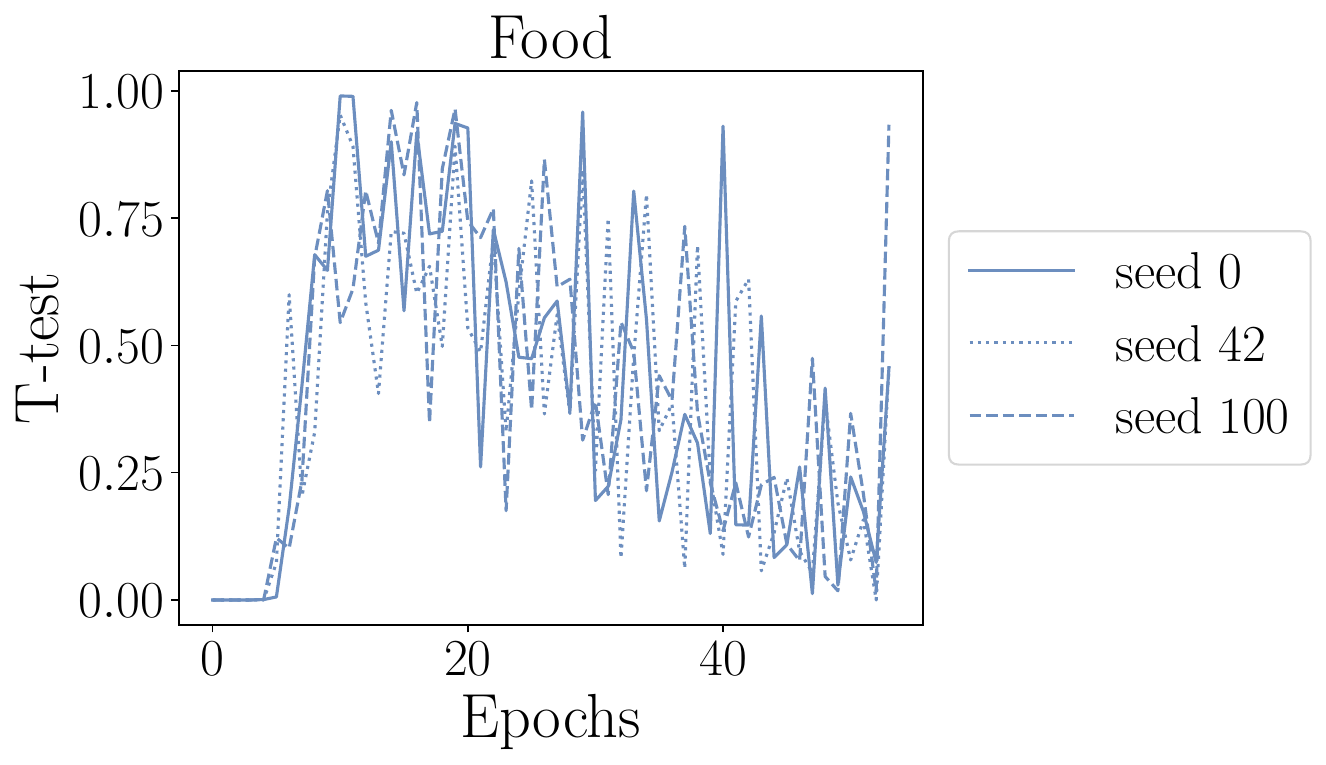}
        \label{fig:subimfood_ttest_mcunet}
    \end{subfigure}
    
    \caption{Evolution of channel gradient norm T-test over time, MobileNetV2 fine-tuned on 3 downstream tasks.}
    \label{fig:ttest_over_time}
\end{figure*}

\subsection{Preliminary Experiments}
\label{sec:prel_exp}
\begin{wrapfigure}{b}{0.5\textwidth}
    \centering
    \includegraphics[width=\linewidth,trim={0 0 0 0},clip]{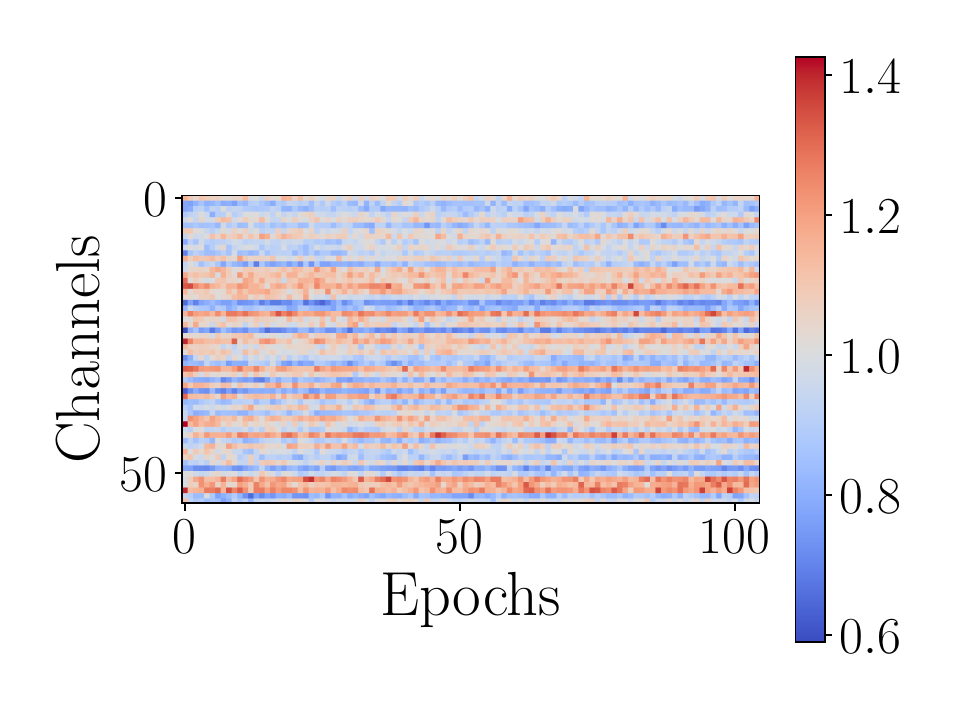}
    \caption{Evolution of channel gradient norm over time within a specific layer of a MobileNetV2 fine-tuned on CIFAR-10. Results are normalized per epoch for visualization.}
    \label{fig:channel_norm_evol}
\end{wrapfigure}
\textbf{LaRa-based layer ranking consistency.} We begin by validating that our proposed LaRa metric introduced in Sec.~\ref{sec:alt_layer_ranking} preserves the architectural consistency of layer importance across downstream tasks. We define layer ranking as the ordered sequence of layers based on their LaRa values, computed according to~\eqref{eq:layer_importance}. For each fine-tuning configuration, we accumulate the LaRa values of each layer over training epochs, forming a "layer topology" vector that characterizes the relative importance distribution across the network architecture. To assess the stability of these rankings across tasks, we calculate Spearman rank correlation coefficients between all pairwise combinations of fine-tuning experiments across our seven evaluation datasets, with three random initializations per dataset. This analysis produces a $21\times 21$ correlation matrix for each architecture, illustrated in Fig.~\ref{fig:layer_sim}.\\
The results confirm that our LaRa metric preserves the architectural consistency property established in Quélennec~\emph{et~al.}'s work. Across all three network architectures, correlation coefficients between downstream task pairs consistently exceed 0.8, with the majority surpassing 0.9. These findings indicate that our LaRa metric maintains the fundamental property that layer importance rankings remain largely consistent across diverse downstream tasks, despite differences in data characteristics and task objectives. This consistency validates the key insight from Quélennec~\emph{et~al.} that layer ranking can be performed \emph{a priori} by conducting a preliminary fine-tuning run on available downstream task data (different than the target downstream task) and collecting the corresponding LaRa metrics. Similar patterns are observed for transformer architectures, as detailed in the appendix.
\begin{figure}[t]
    \centering
    \begin{subfigure}{0.49\columnwidth}
        \includegraphics[width=\linewidth,trim={0 0 0 0},clip]{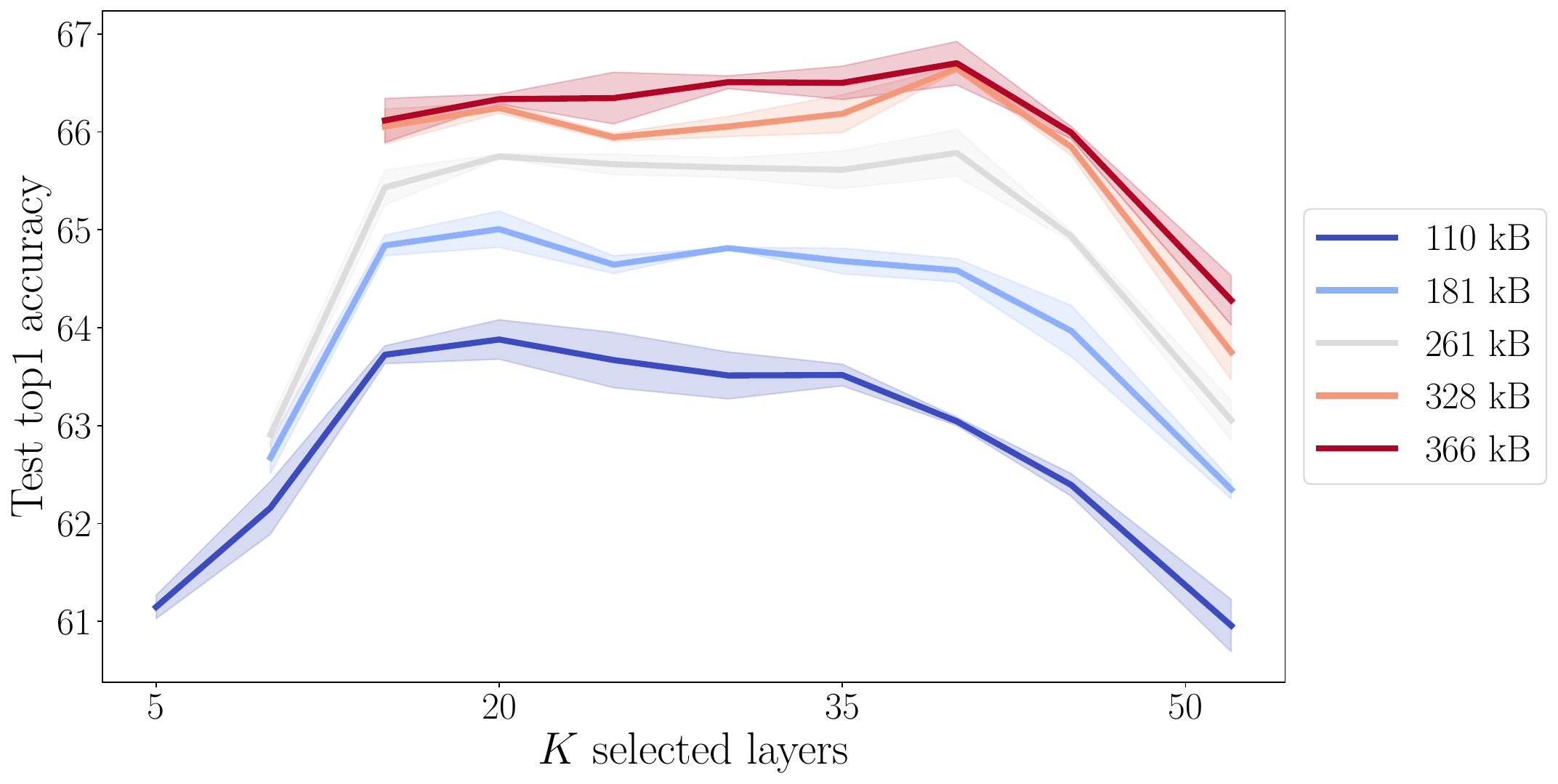}
        \subcaption[]{Final MeDyate test top1 accuracies depending on the number of top $K$ layers for different memory budgets.}
        \label{fig:Med_test_alpha}
    \end{subfigure}
    \begin{subfigure}{0.49\columnwidth}
        \includegraphics[width=\linewidth,trim={0 0 0 0},clip]{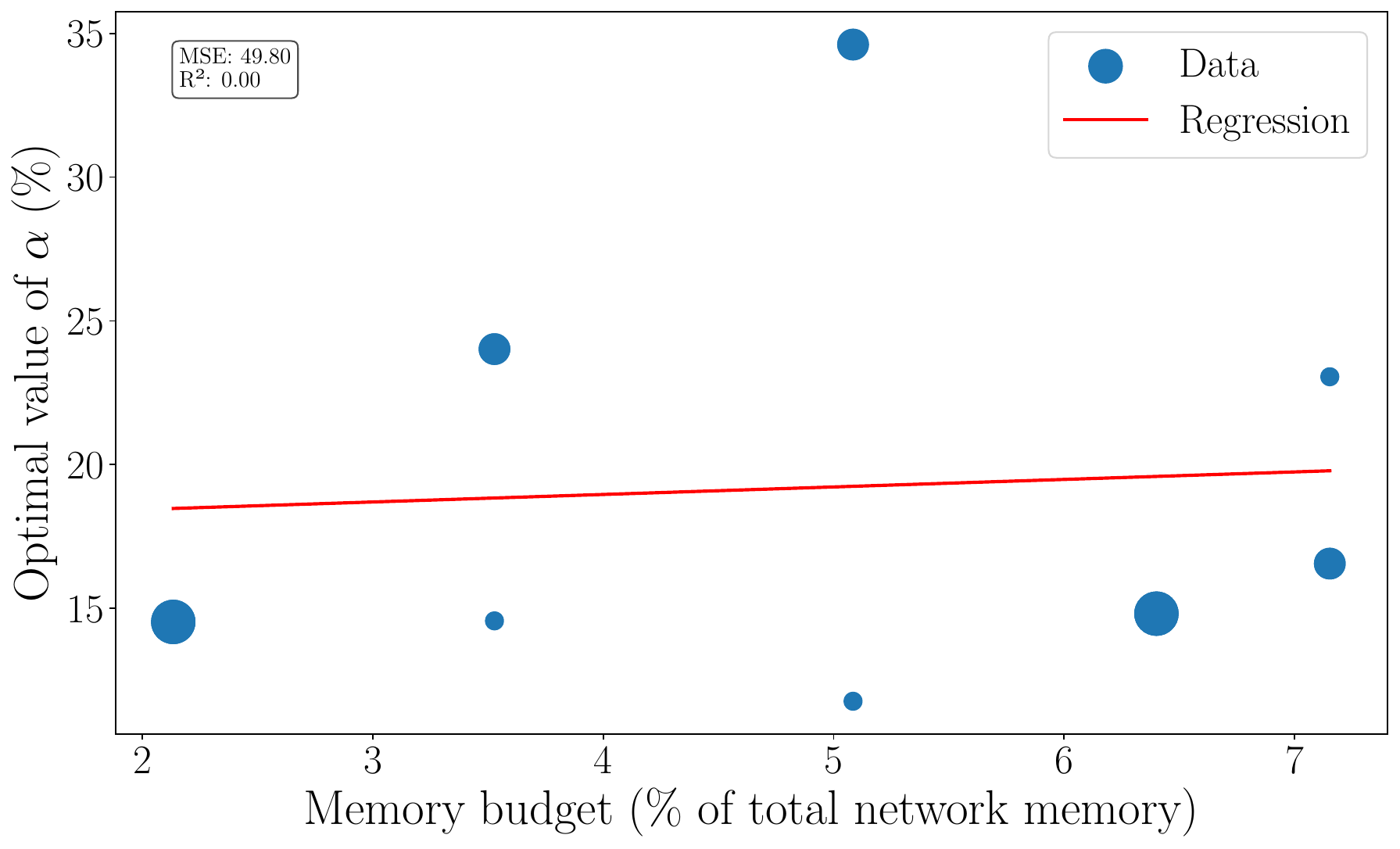}
        \subcaption[]{Regression on the optimal $\alpha_{\text{opt}}$ obtained for different memory budgets and seeds.}
        \label{fig:rel_alpha_budg}
    \end{subfigure}
    \caption{Study of the relationship between $\alpha_{\text{opt}}$ and memory budget. MobileNetV2 fine-tuned on Food.}
    \label{fig:alpha_study}
    \vspace{-15pt}
\end{figure}

\textbf{Channel Gradient Norm Stability.} Building on the theoretical analysis presented in Sec.~\ref{sec:channel_behavior}, we empirically examine the temporal evolution of channel topology during transfer learning across different downstream tasks. Fig.~\ref{fig:ttest_over_time} illustrates this evolution by presenting T-test results comparing channel gradient norms between consecutive epochs to assess distributional similarity over time. The results reveal a clear two-phase pattern: during initial epochs, p-values are consistently equal to zero, indicating significant changes in gradient norm distributions as the network adapts to the new task. Following this initial adaptation period, p-values become substantially higher, failing to reject the topology similarity hypothesis and confirming distributional stabilization. This pattern demonstrates that after the stabilization phase, channels maintaining high gradient norms relative to others preserve this ranking consistently, while channels with comparatively low norms similarly maintain their relative positions.\\
To further illustrate the temporal consistency of relative channel importance, Fig.~\ref{fig:channel_norm_evol} depicts the evolution of channel gradient norms within a specific layer during MobileNetV2 fine-tuning on CIFAR-10, with norms normalized per epoch for visualization clarity. The figure reveals distinct trajectories of consistently high and low gradient norm channels across epochs, forming stable "beams" that persist throughout training. This visualization confirms that the relative gradient norm rankings of channels remain stable over time, supporting the theoretical foundations established in Sec.~\ref{sec:channel_behavior}.\\
\vspace{-1em}

\begin{wrapfigure}{t}{0.5\textwidth}
    \centering
    \includegraphics[width=\linewidth,trim={0 0 0 0},clip]{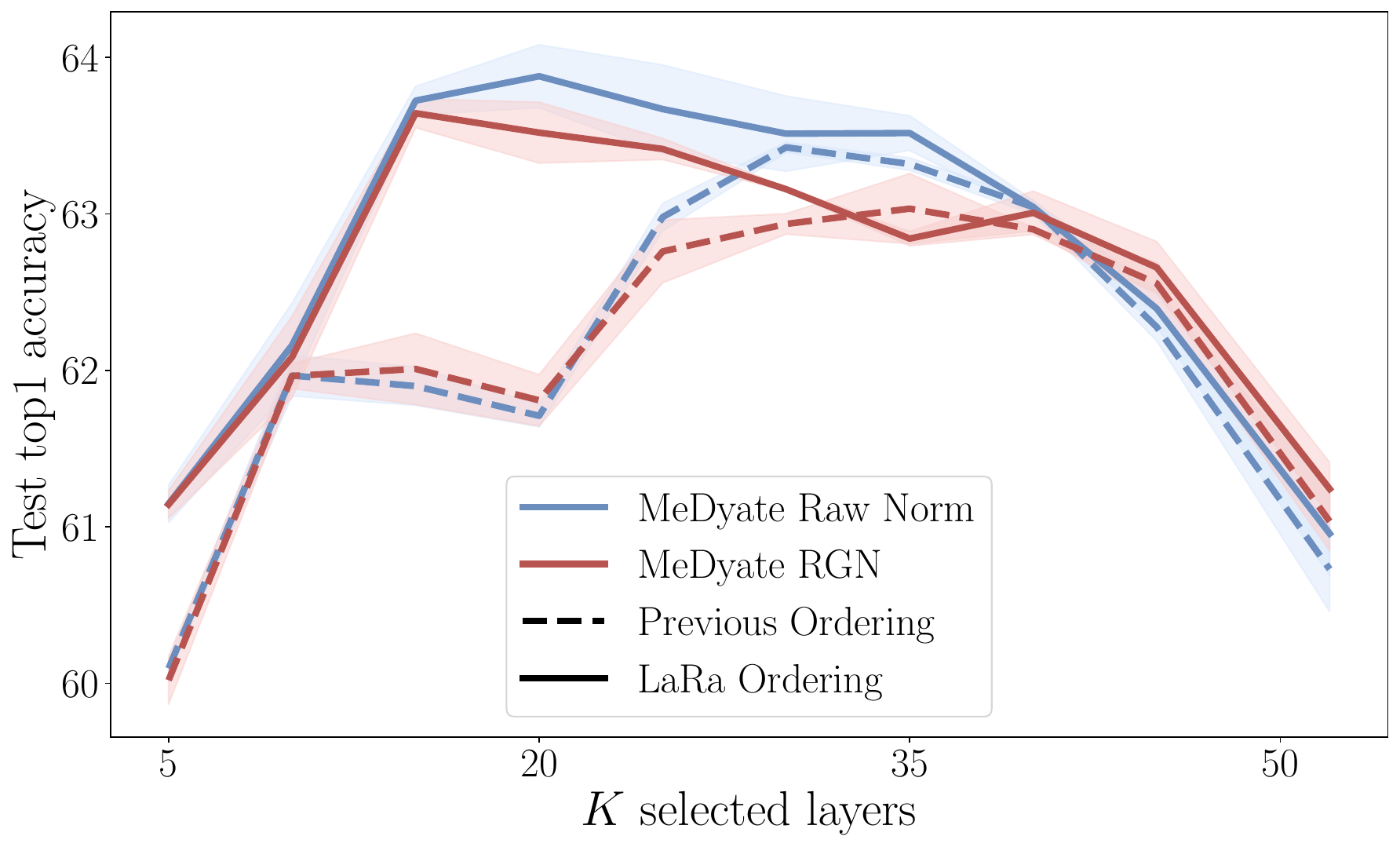}
    \caption{Performance comparison of layer ranking methods (LaRa vs. previous TraDy ordering) and channel importance metrics (raw gradient norm vs. RGN) when fine-tuning MobileNetV2 on Food dataset under memory constraints. Results show test top-1 accuracy across different numbers of selected layers $K$.}
    \label{fig:old_vs_new_raw_vs_RGN}
\end{wrapfigure}
\textbf{Defining $\alpha_{\text{opt}}$.} We investigate the relationship between memory budget constraints and the optimal $\alpha_K$ value that maximizes MeDyate performance. Fig.~\ref{fig:Med_test_alpha} presents the evolution of final test top-1 accuracy as a function of selected layer count $K$ when fine-tuning MobileNetV2 on the Food dataset across different memory budgets. The curves demonstrate that peak accuracy positions shift rightward as budget increases, confirming our hypothesis that optimal search space size correlates with memory constraint as presented in Sec.~\ref{sec:stochasticity_algorithm}. Furthermore, performance typically degrades when layer selection is either too restrictive or too permissive: insufficient layers exclude potentially crucial parameters, while excessive layer inclusion dilutes the memory budget and impairs MeDyate's convergence capacity.\\
To quantify the relationship between $B_{\text{mem}}$ and $\alpha_{\text{opt}}$, we extract the accuracy-maximizing layer count $K$ for each seed-budget combination from Fig.~\ref{fig:Med_test_alpha}, then compute the corresponding $\alpha_K$ values using Eq.~\eqref{eq:layer_hyperparam}. Fig.~\ref{fig:rel_alpha_budg} presents these $\alpha_K$ values as a scatter plot against memory budgets (expressed as percentages of total network memory), with point sizes indicating the frequency of each configuration. Although regression analysis yields high variance, Fig.~\ref{fig:Med_test_alpha} reveals that larger memory budgets achieve relatively stable peak accuracy across a moderate range of $K$ values, providing flexibility in $\alpha_{\text{opt}}$ selection.\\
Given the accuracy degradation observed with insufficient layer selection (corresponding to larger $\alpha_K$ values), we establish $\alpha_{\text{opt}} = 0.2$, positioned slightly above the regression trend. This choice accounts for the discrete nature of layer selection, where we choose $K$ such that $\alpha_K \leq \alpha_{\text{opt}} < \alpha_{K-1}$, ensuring adaptive layer selection that optimizes MeDyate performance across diverse memory constraints. Notably, while Quélennec~\emph{et~al.}'s TraDy achieved peak accuracy with 35 layers under the smallest budget, our LaRa-based ranking reaches optimal performance with only 20 layers, demonstrating the effectiveness of our refined layer ranking methodology.

\textbf{Validating LaRa and Raw Gradient Norm.} We empirically validate the two key modifications proposed in our methodology compared to Quélennec~\emph{et~al.}'s work: the LaRa layer ranking introduced in Sec.~\ref{sec:alt_layer_ranking} and the raw gradient norm channel metric discussed in Sec.~\ref{sec:stochasticity_algorithm}. Our evaluation involves fine-tuning MobileNetV2 on the Food dataset under the smallest memory budget constraint, comparing MeDyate performance across four configurations: our LaRa ranking versus the original TraDy layer ordering, each paired with either raw gradient norm or RGN as the channel importance metric.\\
Fig.~\ref{fig:old_vs_new_raw_vs_RGN} presents the final test top-1 accuracies for each configuration across varying numbers of selected layers $K$. The results demonstrate clear superiority of raw gradient norm over RGN across both layer ranking approaches, supporting our theoretical argument that memory cost reweighting becomes redundant when operating within pre-selected efficient layers. Furthermore, LaRa-based layer ranking consistently achieves higher peak accuracies than the previous ordering while requiring fewer layers to reach optimal performance. This dual advantage validates both the effectiveness of our holistic layer characterization approach and its practical benefits for memory-constrained sampling strategies.

\subsection{Main Results}
\label{sec:main_results}
\begin{wrapfigure}{t}{0.4\textwidth}
    \centering
    \includegraphics[width=\linewidth,trim={0 0 0 0},clip]{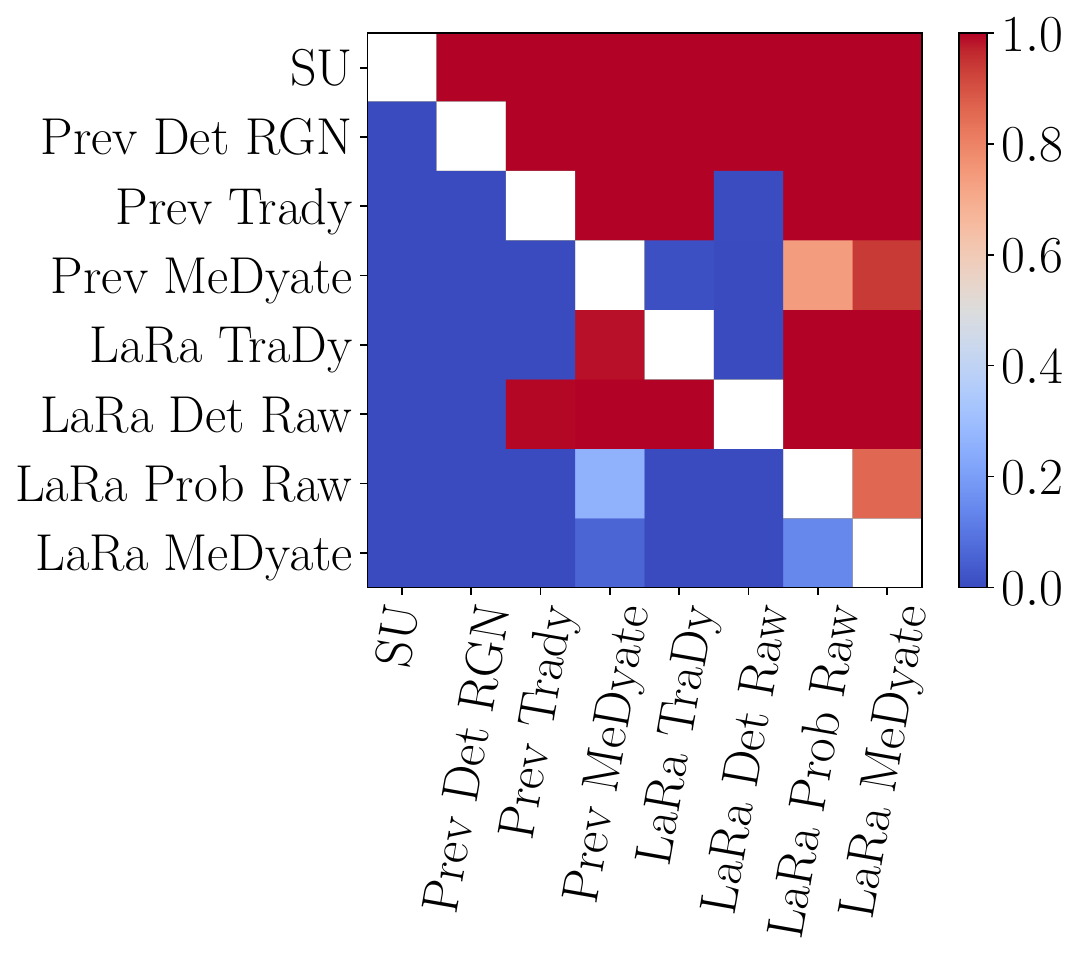}
    \caption{T-test comparisons of average final test accuracies across multiple experimental dimensions.}
    \label{fig:selections_test_ttest_comparison}
    \vspace{-10pt}
\end{wrapfigure}

\textbf{Experimental Design.} We conduct a comprehensive evaluation of our MeDyate strategy across our complete experimental framework. Each channel selection approach is evaluated through 189 individual training runs, representing the full cross-product of three network architectures, seven downstream datasets, three memory budget levels, and three random seeds. Our statistical analysis employs paired T-tests to compare average final test top-1 accuracies across all experimental conditions, as visualized in Fig.~\ref{fig:selections_test_ttest_comparison}. Each matrix cell represents a statistical hypothesis test examining whether the row strategy achieves superior mean accuracy compared to the column strategy. Complete numerical results and additional evaluations on transformer architectures are provided in the appendix (Sec.~\ref{sec:appendix}).

\textbf{Evaluated Strategies.} Our comparative analysis encompasses both static and dynamic selection approaches. We include Lin~\emph{et~al.}'s SU method as the representative static baseline, while all other evaluated strategies employ dynamic channel resampling between epochs, consistent with the demonstrated superiority of dynamic approaches.\\
Our nomenclature distinguishes between layer ranking methodologies: strategies prefixed with \textit{Prev} utilize the original TraDy layer ordering with fixed layer counts, while \textit{LaRa}-prefixed approaches employ our proposed layer ranking with budget-adaptive layer selection. We evaluate several key strategies from the TraDy framework, including deterministic RGN selection (\textit{Det}), which leverages gradient pre-computation to select channels maximizing RGN within memory constraints, and the original TraDy approach for random channel sampling within pre-selected layers.\\
To isolate the contributions of our methodological components, we implement cross-combinations of layer ranking approaches with different channel selection strategies. This includes deterministic raw gradient norm selection within our LaRa framework (\textit{Raw}), MeDyate applied with previous layer selection method, and TraDy adapted to our layer ranking system. Additionally, we evaluate a probabilistic gradient norm strategy (\textit{Prob}) that pre-computes channel gradient norms and converts them to sampling probabilities, serving as a theoretical upper bound for MeDyate's performance by providing complete gradient knowledge during channel selection.

\textbf{Discussion.} The results presented in Fig.~\ref{fig:selections_test_ttest_comparison} support our methodological design choices. MeDyate achieves the highest performance across all strategies, with results that even seem to exceed those of the probabilistic gradient norm strategy with complete gradient information, indicating effective convergence despite memory constraints. The third-ranked performance of Prev MeDyate demonstrates that the core algorithmic principles remain effective even when combined with suboptimal layer ordering and selection, highlighting the robustness of the approach.\\
The strong performance of LaRa TraDy further validates our layer ranking methodology, showing that improved layer selection can enhance existing dynamic strategies. This approach presents a compelling alternative when computational overhead considerations favor simpler sampling schemes over MeDyate's gradient norm computation requirements. The consistent underperformance of raw gradient norm deterministic selection compared to its probabilistic counterparts reinforces the established advantage of stochastic sampling strategies in memory-constrained environments.

\begin{figure*}[t]
    \begin{subfigure}{0.29\textwidth}
        \includegraphics[width=0.9\linewidth]{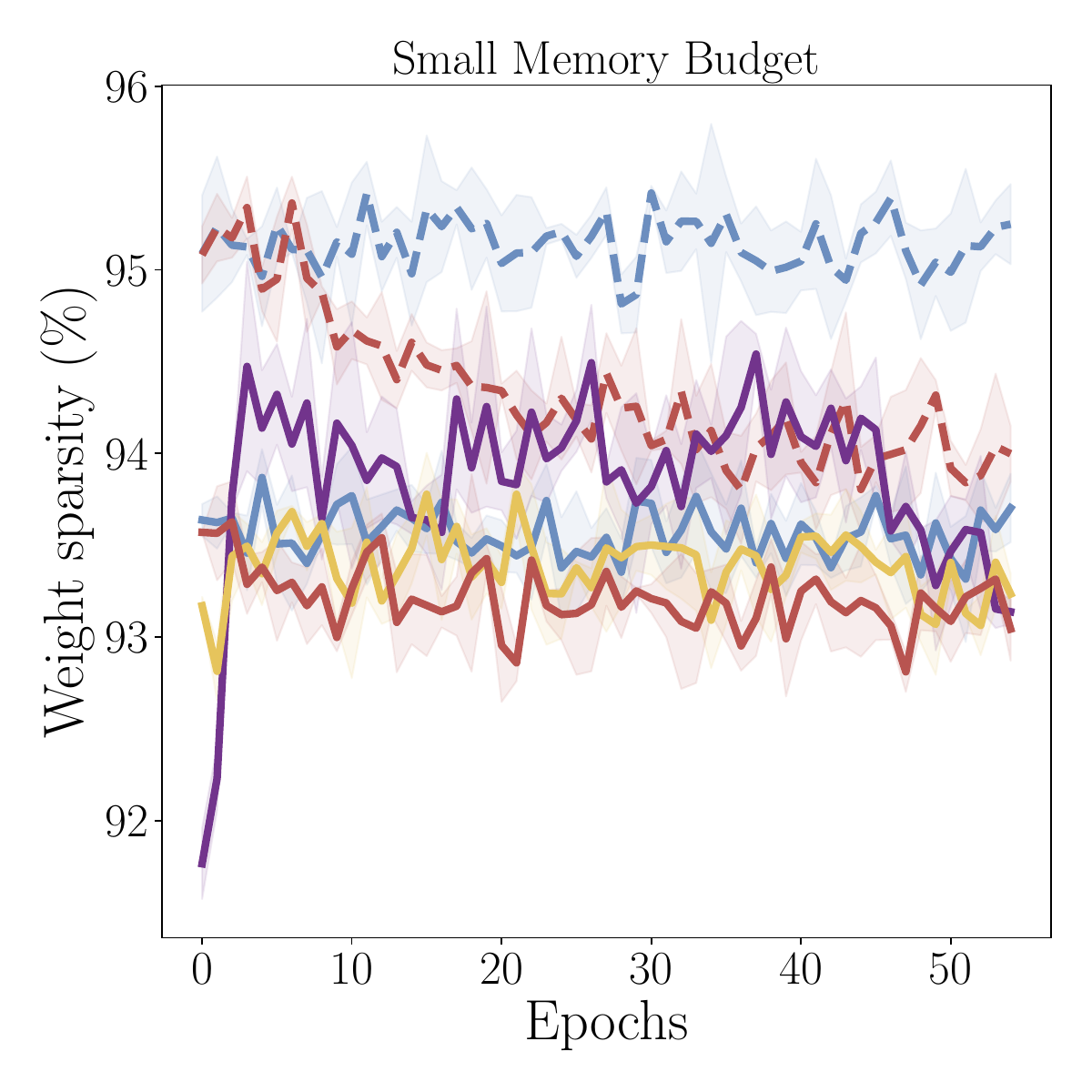}
        \subcaption{Weight sparsity evolution during training.}
        \label{fig:subimfood_weight_spars}
    \end{subfigure}
    \begin{subfigure}{0.29\textwidth}
        \includegraphics[width=0.9\linewidth]{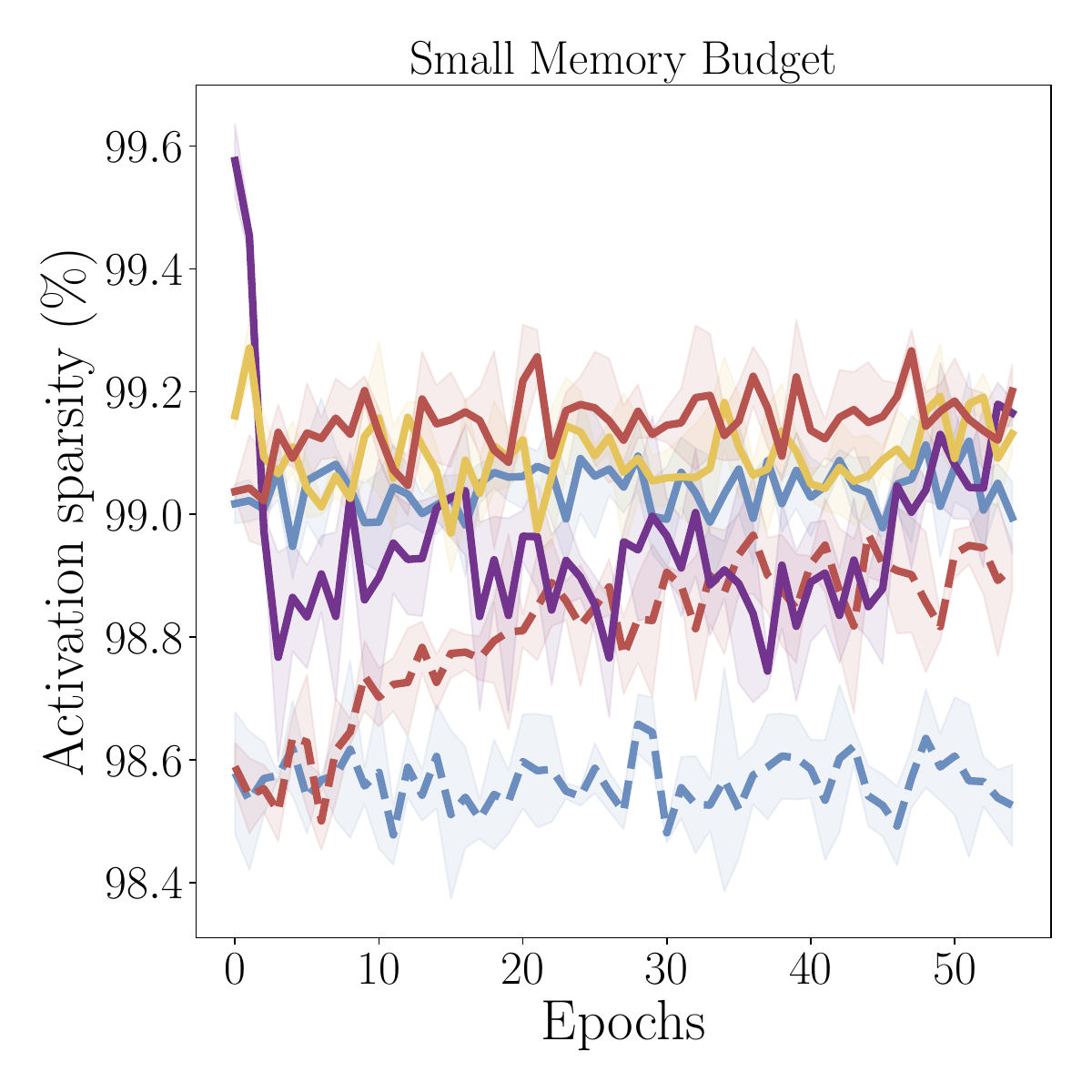}
        \subcaption{Activation sparsity evolution during training.}
        \label{fig:subimfood_act_spars}
    \end{subfigure}
    \begin{subfigure}{0.4\textwidth}
        \includegraphics[width=0.9\linewidth]{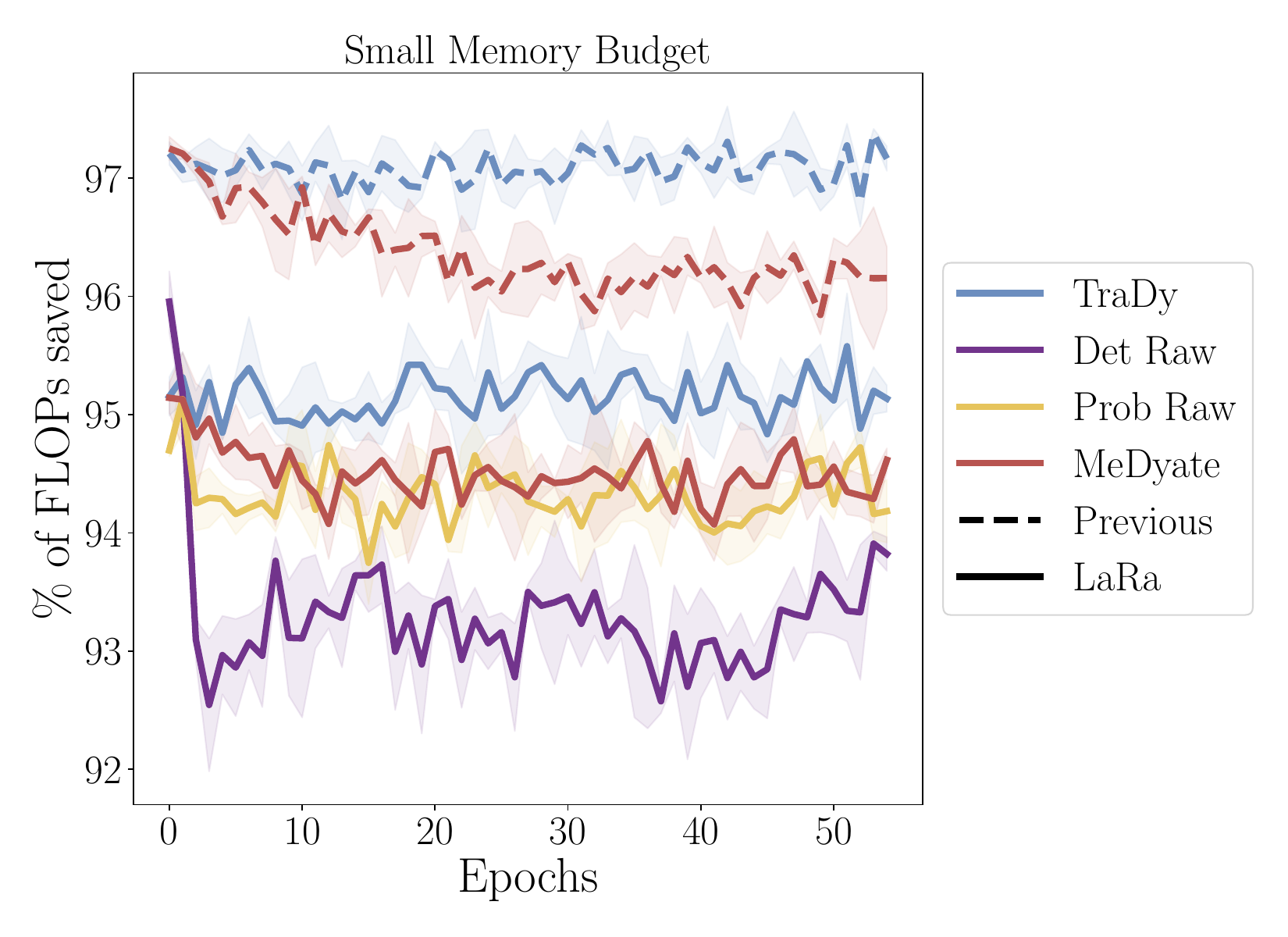}
        \subcaption{Computational savings in weight derivative FLOPs.}
        \label{fig:subimfood_FLOPs}
    \end{subfigure}
    
    \caption{Efficiency metrics comparison across channel selection strategies during MobileNetV2 fine-tuning on Food dataset under memory constraint. Results show evolution of sparsity levels and computational savings throughout training.}
    \label{fig:flops_and_sparsities}
\end{figure*}

\textbf{Efficiency Metrics.} Our algorithm achieves these performance gains while simultaneously reducing computational overhead and maintaining high sparsity levels. Fig.~\ref{fig:flops_and_sparsities} illustrates the evolution of key efficiency metrics when fine-tuning MobileNetV2 on the Food dataset under the smallest memory budget across representative channel selection strategies.\\
The sparsity analysis reveals interesting trade-offs between different approaches. While all methods achieve comparable overall sparsity levels (weight sparsity ranging from 92\% to 96\% and activation sparsity from 98.4\% to 99.6\%), LaRa MeDyate exhibits a distinct pattern of trading lower weight sparsity for higher activation sparsity. This behavior reflects the selection of channels with higher weight-to-activation memory ratios within the LaRa-selected layers. Conversely, TraDy with the previous layer policy demonstrates the opposite tendency, favoring channels with lower weight-to-activation memory ratios.\\
Regarding computational efficiency, Fig.~\ref{fig:subimfood_FLOPs} presents the percentage of weight derivative FLOPs saved through channel freezing during gradient computation (Eq.~\eqref{eq:weight_gradient}). Methods utilizing the previous layer policy achieve higher FLOP savings with typically one to two percentage points more saved FLOPs compared to their LaRa-based counterparts. This difference highlights a potential trade-off between accuracy optimization and computational efficiency, which may be relevant in scenarios where computational cost takes precedence over performance gains.\\
These results represent a single configuration (network, dataset, and memory budget) as metric behaviors remain consistent across experimental conditions. Complete training metrics for all configurations are provided in the supplementary materials.\\

\section{Conclusion}
\label{sec:conclusion}

In this paper, we present MeDyate, a theoretically-grounded framework for memory-constrained dynamic subnetwork adaptation that introduces the LaRa layer ranking metric and exploits channel importance stability during fine-tuning. Our extensive evaluation demonstrates consistent performance improvements over existing approaches while operating within memory budgets as low as a few hundred kB. However, a key limitation of our work is the absence of actual on-device implementation. While our algorithmic innovations show promise in experimental settings, practical deployment of dynamic channel selection on edge devices remains unvalidated, preventing us from obtaining real-world performance metrics. Future research should prioritize efficient on-device implementations to translate these theoretical advances into practical edge AI solutions.

\bibliography{main}
\bibliographystyle{plain}

\appendix
\newpage
\section{Appendix}
\label{sec:appendix}

\subsection{Transformer Results}
\label{sec:appendix_trans_results}

\begin{figure*}[t]
    \begin{subfigure}{0.48\textwidth}
        \includegraphics[width=0.9\linewidth]{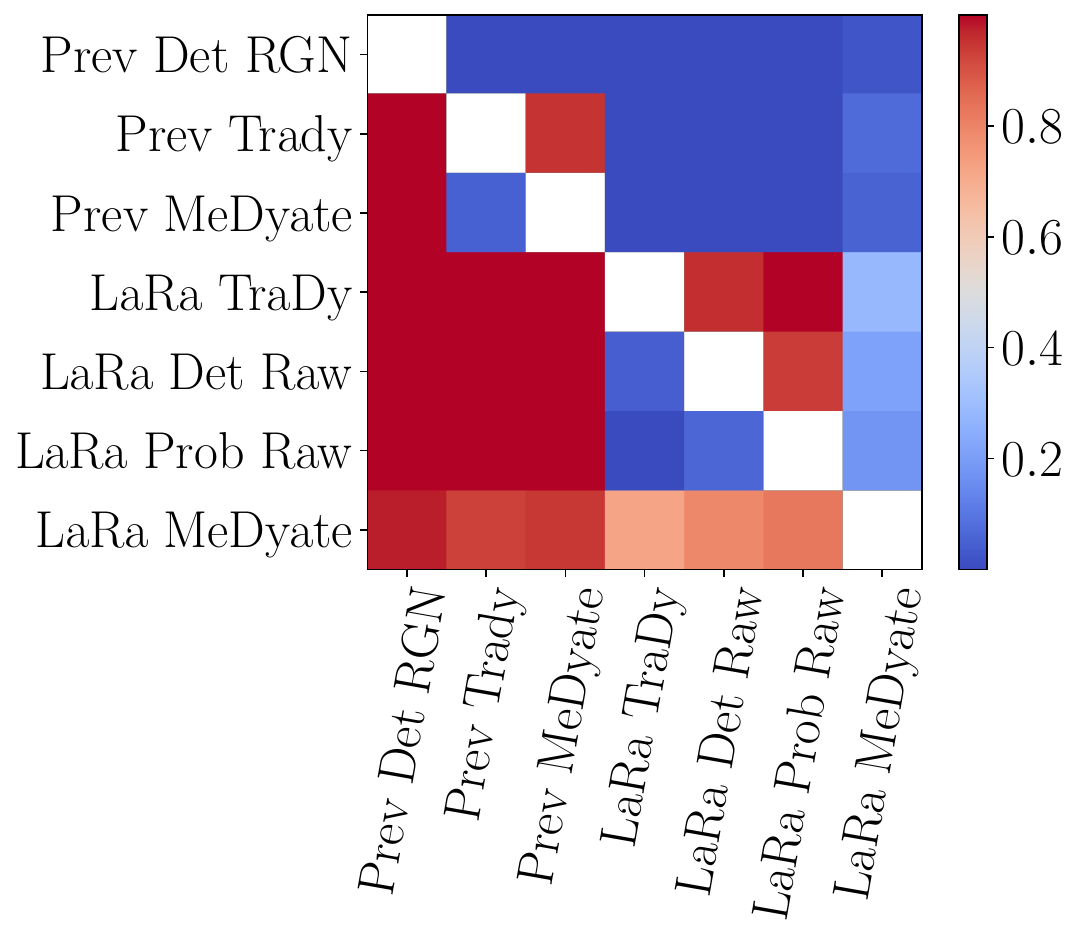}
        \caption{SwinT model}
        \label{fig:subimMeDyate_swint_ttest_final}
    \end{subfigure}
    \begin{subfigure}{0.48\textwidth}
        \includegraphics[width=0.9\linewidth]{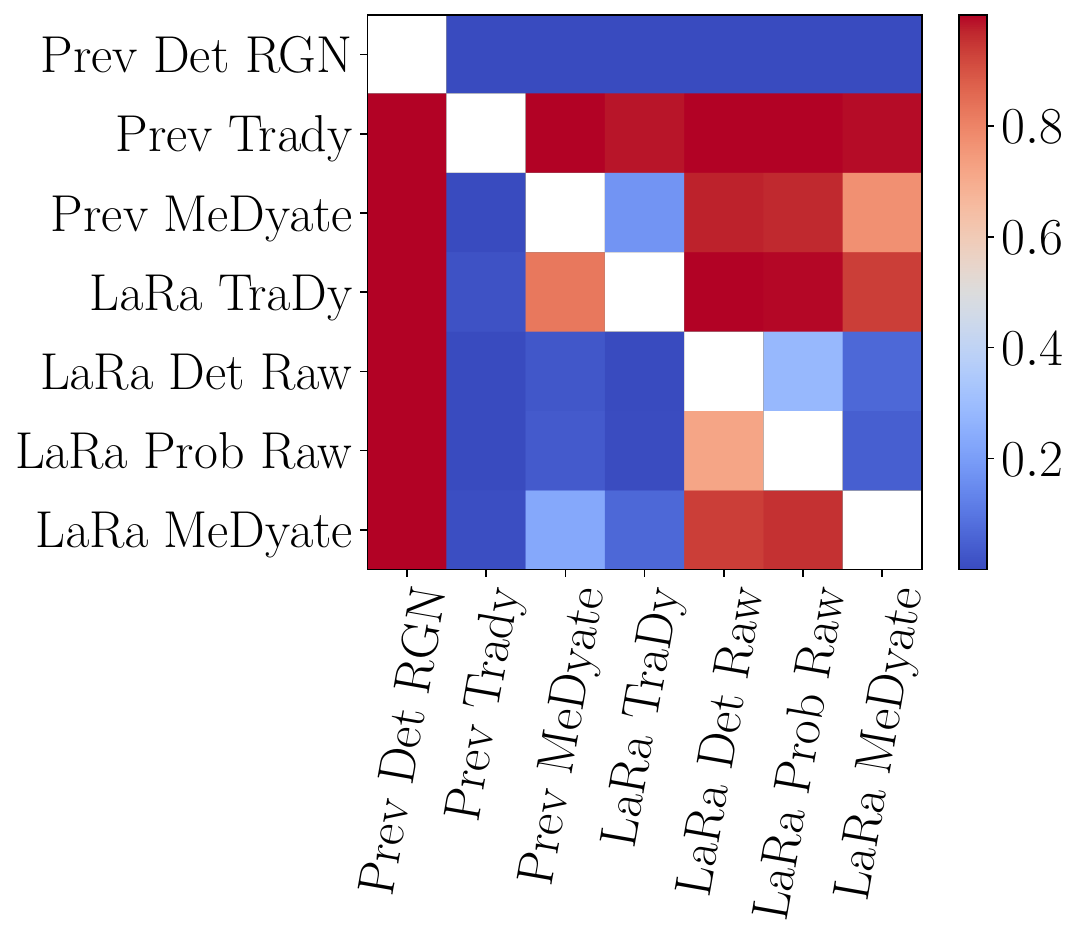}
        \caption{BERTs models}
        \label{fig:subimMeDyate_berts_ttest_final}
    \end{subfigure}
    \caption{T-test comparisons of average final test accuracies across multiple experimental dimensions for each group of transformer architectures.}
    \label{fig:MeDyate_trans_ttest_final}
\end{figure*}

We extend our evaluation to transformer architectures, adopting the framework from~\cite{quélennec2025studytrainingdynamicsmemoryconstrained}. Our experiments employ a SwinT model~\cite{liu2021swin} pre-trained on ImageNet and evaluated on the seven downstream vision tasks, alongside BERT~\cite{kenton2019bert} and RoBERTa~\cite{liu2019roberta} for NLP tasks: QNLI~\cite{demszky2018transforming}, RTE~\cite{poliak2020survey}, and SST2~\cite{socher-etal-2013-recursive}. Fig.~\ref{fig:MeDyate_trans_ttest_final} displays paired T-test comparisons of mean final test accuracies, mirroring the statistical methodology from Sec.~\ref{sec:main_results}. Complete numerical results appear in Sec.~\ref{sec:appendix_tab}.\\
Transformer architectures exhibit significantly different performance patterns compared to CNNs. The deterministic RGN-based channel selection within fixed layer subsets emerges as the dominant strategy across both architecture families, contrasting sharply with CNN results where MeDyate demonstrated clear superiority.\\
The SwinT results (Fig.~\ref{fig:subimMeDyate_swint_ttest_final}) presents an inversion of training dynamics: LaRa-based MeDyate records the lowest performance across all strategies, while the fixed RGN-ranked approach achieves second place and represents the best deployable solution. This inversion indicates that LaRa's design principles, effective for convolutional architectures, introduce counterproductive biases in vision transformers. The adaptive layer selection advantageous for CNNs appears to conflict with the operational characteristics of attention-based vision models.\\
BERTs architectures (Fig.~\ref{fig:subimMeDyate_berts_ttest_final}) show more nuanced behavior. While deterministic RGN selection maintains its lead, performance differences between strategies diminish substantially. LaRa demonstrates acceptable effectiveness in the NLP domain, with deterministic and stochastic raw norm variants achieving competitive accuracy. These findings suggest LaRa's applicability varies with task modality: reasonable for language tasks yet problematic for vision transformers.\\
This architectural divergence reveals fundamental distinctions in fine-tuning behavior requiring systematic analysis. We propose two primary explanatory factors. The attention mechanism's global receptive fields may fundamentally alter gradient propagation patterns relative to the local connectivity assumptions underlying LaRa's formulation for CNNs. Additionally, vision and language transformers process fundamentally different information structures: spatial hierarchies through attention versus sequential token dependencies, potentially necessitating domain-specific layer importance characterization approaches.\\
Future research should explore architecture-aware extensions of the LaRa framework, potentially incorporating attention-specific metrics or developing task-conditional layer ranking strategies tailored to the unique computational patterns of transformer models.

\subsection{Inconclusive Strategy: Weighted Sampling}
\label{sec:appendix_boosting}

\begin{figure}[t]
    \centering
    \includegraphics[width=0.95\linewidth]{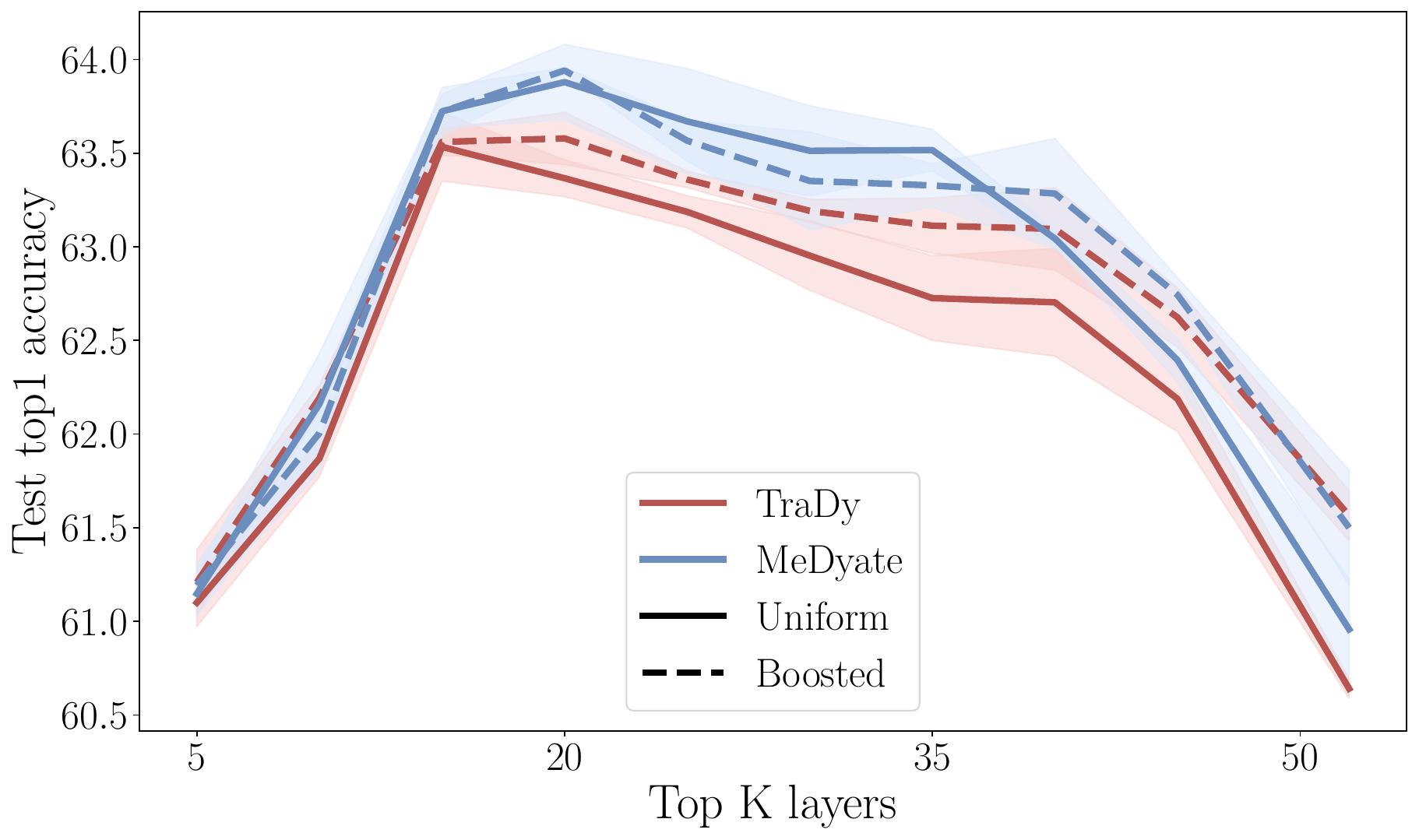}
    \caption{Effects of probability distribution in probabilistic sampling approaches with respect to the number of layers selected when fine-tuning MobileNetV2 on Food dataset under memory constraints. Results show test top-1 accuracy across different numbers of selected layers $K$.}
    \label{fig:uniform_vs_boosted}
\end{figure}

Both Quélennec~\emph{et~al.}'s TRaDy and our proposed MeDyate (Sec.~\ref{sec:stochasticity_algorithm}) employ uniform probability distributions for channel sampling, either throughout the entire fine-tuning process or at initialization. This approach implicitly assigns equal importance to all channels within the selected layer subset. Given our access to pre-computed layer importance through the LaRa metric (Sec.~\ref{sec:alt_layer_ranking}), we investigated replacing uniform sampling with a LaRa-weighted distribution, where channel selection probabilities are proportional to their corresponding layer's LaRa value. We hypothesized that this \textit{boosting} strategy would enable more informed channel selection in TRaDy and accelerate convergence toward the optimal channel distribution in MeDyate.\\
Fig.~\ref{fig:uniform_vs_boosted} illustrates the evolution of final test top-1 accuracy across different numbers of selected LaRa-ranked layers $K$, comparing uniform and boosted variants of both TRaDy and MeDyate. The boosting mechanism primarily influences performance at larger $K$ values, where it mitigates accuracy degradation. However, within the optimal $K$ range that maximizes accuracy, boosting yields only marginal improvements for TRaDy and no statistically significant effect for MeDyate. Given these limited gains, we defer comprehensive exploration of layer-importance-weighted sampling to future work, focusing instead on the more impactful contributions presented in the main paper.

\subsection{Full Results Tables}
\label{sec:appendix_tab}

This section presents experimental results for the selection strategies evaluated in our study. Results from strategies presented in Quélennec~\emph{et~al.}'s work are omitted since we use the same framework to run experiments thus obtaining the same results~\cite{quélennec2025studytrainingdynamicsmemoryconstrained}. Tab.~\ref{tab:MeDyate_CNN} displays results for CNN architectures, while Tab.~\ref{tab:MeDyate_swinT} and Tab.~\ref{tab:MeDyate_NLP} respectively presents results with SwinT and NLP models.

\begin{table*}
    \caption{Comparison of final top1 test accuracies between channel selection strategies over various pretrained convolutional architectures, datasets, and budgets.
    }
    \centering
    \renewcommand*{\arraystretch}{1.2}
    \resizebox{\textwidth}{!}{
    \label{tab:MeDyate_CNN}
        \begin{tabular}{ccccccccccc}
            \toprule
            \textbf{Model} & $B_{\text{mem}}$ &\textbf{Method} & \textbf{CIFAR-10} & \textbf{CIFAR-100} & \textbf{CUB} & \textbf{Flowers}  & \textbf{Food} & \textbf{Pets} & \textbf{VWW} & \textbf{Average} \\
            
            \midrule
            \multirow{20}{*}{MbV2-w0.35} 
            & \multirow{6}{*}{27 946} & Prev MeDyate &  90.18\small$\pm$0.13 & 68.84\small$\pm$0.20 & 58.19\small$\pm$0.64 & 80.07\small$\pm$0.42 & 63.32\small$\pm$0.04 & 77.02\small$\pm$0.26 & 88.66\small$\pm$0.11 & 75.18\small$\pm$0.85 \\

            \cmidrule{3-11}
            & & LaRa TraDy & 89.97\small$\pm$0.06 & 68.67\small$\pm$0.23 & 58.67\small$\pm$0.26 & 80.34\small$\pm$0.60 & 63.60\small$\pm$0.12 & 77.72\small$\pm$0.24 & 88.48\small$\pm$0.25 & 75.35\small$\pm$0.79 \\

            \cmidrule{3-11}
            & & LaRa Det Raw & 89.89\small$\pm$0.03 & 68.17\small$\pm$0.20 & 58.0\small$\pm$0.19 & 80.65\small$\pm$0.35 & 62.61\small$\pm$0.01 & 77.60\small$\pm$0.45 & 88.10\small$\pm$0.29 & 75.00\small$\pm$0.70 \\

            \cmidrule{3-11}
            & & LaRa Prob Raw & 90.13\small$\pm$0.17 & 68.75\small$\pm$0.23 & 58.20\small$\pm$0.36 & 80.25\small$\pm$0.42 & 63.93\small$\pm$0.07 & 77.28\small$\pm$0.20 & 88.58\small$\pm$0.31 & 75.30\small$\pm$0.73 \\

            \cmidrule{3-11}
            & & LaRa MeDyate & 90.24\small$\pm$0.11 & 68.83\small$\pm$0.13 & 58.73\small$\pm$0.12 & 80.31\small$\pm$0.15 & 63.7\small$\pm$0.16 & 77.45\small$\pm$0.25 & 88.53\small$\pm$0.11 & 75.40\small$\pm$0.41 \\

            \cmidrule{2-11}
            & \multirow{6}{*}{66 592} & Prev MeDyate & 90.70\small$\pm$0.05 & 69.66\small$\pm$0.17 & 58.96\small$\pm$0.33 & 80.80\small$\pm$0.34 & 65.78\small$\pm$0.11 & 77.16\small$\pm$0.70 & 87.65\small$\pm$0.14 & 75.82\small$\pm$0.88 \\

            \cmidrule{3-11}
            & & LaRa TraDy & 90.82\small$\pm$0.18 & 69.65\small$\pm$0.16 & 58.50\small$\pm$0.23 & 80.27\small$\pm$0.29 & 65.08\small$\pm$0.22 & 76.90\small$\pm$0.20 & 88.06\small$\pm$0.22 & 75.61\small$\pm$0.58 \\

            \cmidrule{3-11}
            & & LaRa Det Raw & 90.57\small$\pm$0.12 & 69.22\small$\pm$0.16 & 58.62\small$\pm$0.08 & 81.10\small$\pm$0.62 & 64.58\small$\pm$0.26 & 77.24\small$\pm$0.38 & 87.98\small$\pm$0.05 & 75.62\small$\pm$0.80 \\

            \cmidrule{3-11}
            & & LaRa Prob Raw & 91.02\small$\pm$0.09 & 69.59\small$\pm$0.28 & 58.87\small$\pm$0.24 & 80.68\small$\pm$0.51 & 65.38\small$\pm$0.25 & 77.00\small$\pm$0.59 & 87.72\small$\pm$0.24 & 75.75\small$\pm$0.93 \\

            \cmidrule{3-11}
            & & LaRa MeDyate & 91.02\small$\pm$0.18 & 69.70\small$\pm$0.31 & 58.90\small$\pm$0.10 & 80.71\small$\pm$0.59 & 65.37\small$\pm$0.08 & 76.92\small$\pm$0.10 & 88.01\small$\pm$0.29 & 75.80\small$\pm$0.77 \\

            \cmidrule{2-11}
            & \multirow{6}{*}{93 696} & Prev MeDyate & 91.27\small$\pm$0.11 & 70.23\small$\pm$0.31 & 58.86\small$\pm$0.54 & 81.27\small$\pm$0.43 & 66.72\small$\pm$0.19 & 77.11\small$\pm$0.33 & 88.02\small$\pm$0.18 & 76.21\small$\pm$0.87 \\

            \cmidrule{3-11}
            & & LaRa TraDy & 91.37\small$\pm$0.06 & 70.09\small$\pm$0.21 & 59.03\small$\pm$0.32 & 80.66\small$\pm$0.32 & 66.39\small$\pm$0.06 & 76.86\small$\pm$0.24 & 87.96\small$\pm$0.21 & 76.05\small$\pm$0.60 \\

            \cmidrule{3-11}
            & & LaRa Det Raw & 90.91\small$\pm$0.04 & 69.73\small$\pm$0.08 & 58.88\small$\pm$0.24 & 81.48\small$\pm$0.35 & 65.53\small$\pm$0.12 & 77.24\small$\pm$0.46 & 87.93\small$\pm$0.26 & 75.96\small$\pm$0.69 \\

            \cmidrule{3-11}
            & & LaRa Prob Raw & 91.44\small$\pm$0.10 & 70.23\small$\pm$0.15 & 59.45\small$\pm$0.03 & 81.05\small$\pm$0.39 & 66.83\small$\pm$0.04 & 77.13\small$\pm$0.14 & 87.87\small$\pm$0.19 & 76.29\small$\pm$0.49 \\

            \cmidrule{3-11}
            & & LaRa MeDyate & 91.35\small$\pm$0.15 & 70.21\small$\pm$0.05 & 59.08\small$\pm$0.36 & 80.94\small$\pm$0.40 & 66.99\small$\pm$0.14 & 77.04\small$\pm$0.08 & 87.9\small$\pm$0.36 & 76.22\small$\pm$0.69 \\

            \cmidrule{2-11}
            & 1 252 320 & Baseline & 92.72\small$\pm$0.03 & 72.69\small$\pm$0.16 & 60.03\small$\pm$0.18 & 81.88\small$\pm$0.34 & 70.79\small$\pm$0.20 & 76.68\small$\pm$0.33 & 88.58\small$\pm$0.19 & 77.62\small$\pm$0.60 \\

            \midrule
            \multirow{20}{*}{MCUNet-in1} 
            & \multirow{6}{*}{15 936} & Prev MeDyate & 90.63\small$\pm$0.10 & 69.94\small$\pm$0.12 & 61.70\small$\pm$0.05 & 82.51\small$\pm$0.22 & 66.94\small$\pm$0.23 & 81.21\small$\pm$0.04 & 89.69\small$\pm$0.33 & 77.52\small$\pm$0.49 \\

            \cmidrule{3-11}
            & & LaRa TraDy & 90.89\small$\pm$0.06 & 69.88\small$\pm$0.13 & 62.23\small$\pm$0.07 & 83.00\small$\pm$0.41 & 67.34\small$\pm$0.15 & 81.62\small$\pm$0.24 & 89.72\small$\pm$0.24 & 77.81\small$\pm$0.58 \\

            \cmidrule{3-11}
            & & LaRa Det Raw & 90.28\small$\pm$0.14 & 69.66\small$\pm$0.28 & 61.83\small$\pm$0.63 & 82.65\small$\pm$0.40 & 66.29\small$\pm$0.19 & 80.85\small$\pm$0.34 & 89.67\small$\pm$0.15 & 77.32\small$\pm$0.91 \\

            \cmidrule{3-11}
            & & LaRa Prob Raw & 90.91\small$\pm$0.18 & 69.74\small$\pm$0.20 & 61.75\small$\pm$0.39 & 82.61\small$\pm$0.19 & 67.33\small$\pm$0.26 & 80.98\small$\pm$0.26 & 89.55\small$\pm$0.21 & 77.55\small$\pm$0.66 \\

            \cmidrule{3-11}
            & & LaRa MeDyate & 90.89\small$\pm$0.10 & 69.89\small$\pm$0.05 & 62.39\small$\pm$0.41 & 82.76\small$\pm$0.43 & 67.39\small$\pm$0.12 & 81.37\small$\pm$0.24 & 89.94\small$\pm$0.04 & 77.80\small$\pm$0.66 \\

            \cmidrule{2-11}
            & \multirow{6}{*}{64 832} & Prev MeDyate & 92.33\small$\pm$0.12 & 72.41\small$\pm$0.25 & 62.40\small$\pm$0.50 & 83.29\small$\pm$0.63 & 71.20\small$\pm$0.28 & 81.44\small$\pm$0.37 & 89.22\small$\pm$0.31 & 78.90\small$\pm$1.02 \\

            \cmidrule{3-11}
            & & LaRa TraDy & 92.09\small$\pm$0.02 & 72.24\small$\pm$0.38 & 62.60\small$\pm$0.34 & 83.07\small$\pm$0.53 & 70.44\small$\pm$0.03 & 81.46\small$\pm$0.44 & 89.09\small$\pm$0.24 & 78.71\small$\pm$0.89 \\

            \cmidrule{3-11}
            & & LaRa Det Raw & 91.94\small$\pm$0.30 & 71.57\small$\pm$0.11 & 62.29\small$\pm$0.50 & 82.71\small$\pm$0.40 & 69.98\small$\pm$0.16 & 80.96\small$\pm$0.09 & 89.19\small$\pm$0.15 & 78.38\small$\pm$0.75 \\

            \cmidrule{3-11}
            & & LaRa Prob Raw & 92.14\small$\pm$0.10 & 72.26\small$\pm$0.18 & 62.51\small$\pm$0.29 & 82.93\small$\pm$0.42 & 70.79\small$\pm$0.10 & 81.59\small$\pm$0.45 & 89.32\small$\pm$0.29 & 78.79\small$\pm$0.77 \\

            \cmidrule{3-11}
            & & LaRa MeDyate & 92.23\small$\pm$0.18 & 72.00\small$\pm$0.13 & 62.44\small$\pm$0.09 & 83.24\small$\pm$0.29 & 71.00\small$\pm$0.19 & 81.20\small$\pm$0.03 & 89.18\small$\pm$0.23 & 78.76\small$\pm$0.48 \\

            \cmidrule{2-11}
            & \multirow{6}{*}{112 640} & Prev MeDyate & 92.78\small$\pm$0.09 & 72.88\small$\pm$0.36 & 61.83\small$\pm$0.49 & 83.48\small$\pm$0.44 & 72.42\small$\pm$0.29 & 81.16\small$\pm$0.48 & 89.14\small$\pm$0.49 & 79.10\small$\pm$1.06 \\

            \cmidrule{3-11}
            & & LaRa TraDy & 92.78\small$\pm$0.10 & 73.71\small$\pm$0.03 & 61.85\small$\pm$0.54 & 83.24\small$\pm$0.26 & 72.20\small$\pm$0.14 & 81.17\small$\pm$0.55 & 89.14\small$\pm$0.09 & 79.16\small$\pm$0.84 \\

            \cmidrule{3-11}
            & & LaRa Det Raw & 92.51\small$\pm$0.19 & 72.88\small$\pm$0.20 & 62.14\small$\pm$0.12 & 83.16\small$\pm$0.34 & 71.68\small$\pm$0.11 & 80.78\small$\pm$0.28 & 89.59\small$\pm$0.29 & 78.96\small$\pm$0.62 \\

            \cmidrule{3-11}
            & & LaRa Prob Raw & 92.78\small$\pm$0.23 & 73.46\small$\pm$0.21 & 62.06\small$\pm$0.30 & 82.96\small$\pm$0.23 & 72.55\small$\pm$0.11 & 81.15\small$\pm$0.19 & 89.50\small$\pm$0.06 & 79.21\small$\pm$0.54 \\

            \cmidrule{3-11}
            & & LaRa MeDyate & 92.72\small$\pm$0.20 & 73.55\small$\pm$0.28 & 62.36\small$\pm$0.34 & 83.46\small$\pm$0.42 & 72.73\small$\pm$0.02 & 80.83\small$\pm$0.50 & 89.51\small$\pm$0.13 & 79.31\small$\pm$0.82 \\

            \cmidrule{2-11}
            & 1 309 808 & Baseline & 93.87\small$\pm$0.10 & 76.03\small$\pm$0.18 & 61.62\small$\pm$0.62 & 83.45\small$\pm$0.42 & 75.74\small$\pm$0.14 & 79.49\small$\pm$0.60 & 90.06\small$\pm$0.16 & 80.04\small$\pm$1.00 \\

            \midrule
            \multirow{20}{*}{Proxyless-w0.3} 
            & \multirow{6}{*}{25 984} & Prev MeDyate & 91.43\small$\pm$0.20 & 70.30\small$\pm$0.08 & 57.74\small$\pm$0.38 & 82.15\small$\pm$0.22 & 64.80\small$\pm$0.24 & 78.77\small$\pm$0.05 & 88.57\small$\pm$0.13 & 76.25\small$\pm$0.56 \\

            \cmidrule{3-11}
            & & LaRa TraDy & 91.46\small$\pm$0.17 & 69.38\small$\pm$0.15 & 58.18\small$\pm$0.30 & 81.76\small$\pm$0.25 & 64.58\small$\pm$0.05 & 79.10\small$\pm$0.22 & 88.57\small$\pm$0.05 & 76.15\small$\pm$0.51 \\

            \cmidrule{3-11}
            & & LaRa Det Raw & 91.26\small$\pm$0.11 & 68.59\small$\pm$0.11 & 57.87\small$\pm$0.10 & 82.11\small$\pm$0.31 & 63.40\small$\pm$0.07 & 78.91\small$\pm$0.16 & 88.65\small$\pm$0.18 & 75.83\small$\pm$0.44 \\

            \cmidrule{3-11}
            & & LaRa Prob Raw & 91.53\small$\pm$0.24 & 69.39\small$\pm$0.12 & 58.14\small$\pm$0.28 & 81.89\small$\pm$0.40 & 64.66\small$\pm$0.24 & 79.27\small$\pm$0.52 & 88.74\small$\pm$0.17 & 76.23\small$\pm$0.82 \\

            \cmidrule{3-11}
            & & LaRa MeDyate & 91.38\small$\pm$0.21 & 69.71\small$\pm$0.10 & 58.27\small$\pm$0.06 & 81.96\small$\pm$0.46 & 64.75\small$\pm$0.27 & 78.92\small$\pm$0.24 & 88.77\small$\pm$0.20 & 76.25\small$\pm$0.66 \\

            \cmidrule{2-11}
            & \multirow{6}{*}{72 960} & Prev MeDyate & 92.27\small$\pm$0.13 & 71.76\small$\pm$0.13 & 59.06\small$\pm$0.29 & 82.59\small$\pm$0.20 & 67.81\small$\pm$0.08 & 79.28\small$\pm$0.35 & 88.46\small$\pm$0.02 & 77.32\small$\pm$0.54 \\

            \cmidrule{3-11}
            & & LaRa TraDy & 92.29\small$\pm$0.24 & 71.26\small$\pm$0.17 & 58.73\small$\pm$0.24 & 82.50\small$\pm$0.33 & 67.14\small$\pm$0.13 & 79.27\small$\pm$0.22 & 88.44\small$\pm$0.26 & 77.09\small$\pm$0.62 \\

            \cmidrule{3-11}
            & & LaRa Det Raw & 92.32\small$\pm$0.07 & 71.32\small$\pm$0.03 & 58.87\small$\pm$0.20 & 83.09\small$\pm$0.28 & 67.20\small$\pm$0.35 & 78.49\small$\pm$0.20 & 88.02\small$\pm$0.24 & 77.04\small$\pm$0.59 \\

            \cmidrule{3-11}
            & & LaRa Prob Raw & 92.48\small$\pm$0.04 & 71.83\small$\pm$0.33 & 59.75\small$\pm$0.67 & 82.89\small$\pm$0.46 & 67.85\small$\pm$0.19 & 79.31\small$\pm$0.47 & 88.02\small$\pm$0.18 & 77.45\small$\pm$1.03 \\

            \cmidrule{3-11}
            & & LaRa MeDyate & 92.50\small$\pm$0.15 & 71.54\small$\pm$0.17 & 59.25\small$\pm$0.79 & 82.90\small$\pm$0.12 & 67.66\small$\pm$0.17 & 79.05\small$\pm$0.32 & 88.05\small$\pm$0.14 & 77.28\small$\pm$0.92 \\

            \cmidrule{2-11}
            & \multirow{6}{*}{101 376} & Prev MeDyate & 92.65\small$\pm$0.25 & 72.28\small$\pm$0.26 & 59.58\small$\pm$0.20 & 83.04\small$\pm$0.30 & 68.77\small$\pm$0.21 & 79.23\small$\pm$0.22 & 88.35\small$\pm$0.17 & 77.70\small$\pm$0.62 \\

            \cmidrule{3-11}
            & & LaRa TraDy & 92.56\small$\pm$0.09 & 71.76\small$\pm$0.29 & 59.58\small$\pm$0.48 & 82.6\small$\pm$0.20 & 67.79\small$\pm$0.16 & 79.05\small$\pm$0.18 & 88.3\small$\pm$0.20 & 77.38\small$\pm$0.68 \\

            \cmidrule{3-11}
            & & LaRa Det Raw & 92.37\small$\pm$0.19 & 71.36\small$\pm$0.31 & 59.34\small$\pm$0.61 & 83.25\small$\pm$0.46 & 67.35\small$\pm$0.07 & 79.00\small$\pm$0.45 & 88.30\small$\pm$0.27 & 77.28\small$\pm$1.00 \\

            \cmidrule{3-11}
            & & LaRa Prob Raw & 92.61\small$\pm$0.15 & 71.62\small$\pm$0.12 & 59.91\small$\pm$0.26 & 82.99\small$\pm$0.28 & 68.32\small$\pm$0.14 & 79.31\small$\pm$0.36 & 88.47\small$\pm$0.45 & 77.60\small$\pm$0.73 \\

            \cmidrule{3-11}
            & & LaRa MeDyate & 92.42\small$\pm$0.02 & 71.86\small$\pm$0.09 & 59.73\small$\pm$0.28 & 82.9\small$\pm$0.23 & 68.52\small$\pm$0.22 & 79.45\small$\pm$0.50 & 88.23\small$\pm$0.14 & 77.59\small$\pm$0.68 \\

            \cmidrule{2-11}
            & 1 162 032 & Baseline & 93.71\small$\pm$0.12 & 74.81\small$\pm$0.13 & 61.75\small$\pm$0.12 & 84.44\small$\pm$0.50 & 72.98\small$\pm$0.09 & 78.53\small$\pm$0.10 &  88.95\small$\pm$0.04 & 79.31\small$\pm$0.56 \\
            \bottomrule
        \end{tabular}
    }
\end{table*}
\begin{table*}
    \caption{Comparison of final top1 test accuracies between channel selection strategies over various datasets, and budgets when considering a SwinT architecture
    }
    \centering
    \renewcommand*{\arraystretch}{1.2}
    \resizebox{\textwidth}{!}{
    \label{tab:MeDyate_swinT}
        \begin{tabular}{ccccccccccc}
            \toprule
            \textbf{Model} & $B_{\text{mem}}$ &\textbf{Method} & \textbf{CIFAR-10} & \textbf{CIFAR-100} & \textbf{CUB} & \textbf{Flowers}  & \textbf{Food} & \textbf{Pets} & \textbf{VWW} & \textbf{Average} \\
            
            \midrule
            \multirow{26}{*}{SwinT} 
            & \multirow{6}{*}{27 946} & Prev MeDyate & 96.35\small$\pm$0.11 & 82.91\small$\pm$0.10 & 73.98\small$\pm$0.06 & 88.44\small$\pm$0.34 & 80.74\small$\pm$0.04 & 90.97\small$\pm$0.20 & 93.75\small$\pm$0.15 & 86.73\small$\pm$0.45 \\

            \cmidrule{3-11}
            & & LaRa TraDy & 96.30\small$\pm$0.07 & 83.16\small$\pm$0.17 & 74.31\small$\pm$0.23 & 88.59\small$\pm$0.24 & 80.78\small$\pm$0.05 & 90.91\small$\pm$0.12 & 92.77\small$\pm$0.14 & 86.69\small$\pm$0.43 \\

            \cmidrule{3-11}
            & & LaRa Det Raw & 96.34\small$\pm$0.06 & 82.96\small$\pm$0.10 & 74.42\small$\pm$0.10 & 88.21\small$\pm$0.14 & 80.76\small$\pm$0.11 & 90.98\small$\pm$0.11 & 92.93\small$\pm$0.14 & 86.66\small$\pm$0.29 \\

            \cmidrule{3-11}
            & & LaRa Prob Raw & 96.33\small$\pm$0.04 & 83.14\small$\pm$0.18 & 74.35\small$\pm$0.02 & 88.56\small$\pm$0.34 & 80.90\small$\pm$0.17 & 90.87\small$\pm$0.29 & 92.81\small$\pm$0.06 & 86.71\small$\pm$0.52 \\

            \cmidrule{3-11}
            & & LaRa MeDyate & 96.29\small$\pm$0.09 & 83.10\small$\pm$0.06 & 74.34\small$\pm$0.01 & 88.63\small$\pm$0.51 & 80.97\small$\pm$0.05 & 90.99\small$\pm$0.19 & 92.76\small$\pm$0.07 & 86.73\small$\pm$0.56 \\

            \cmidrule{2-11}
            & \multirow{6}{*}{112 640} & Prev MeDyate & 96.76\small$\pm$0.05 & 83.58\small$\pm$0.16 & 74.43\small$\pm$0.15 & 88.85\small$\pm$0.23 & 81.58\small$\pm$0.04 & 90.99\small$\pm$0.19 & 93.55\small$\pm$0.38 & 87.11\small$\pm$0.53 \\

            \cmidrule{3-11}
            & & LaRa TraDy & 96.94\small$\pm$0.14 & 83.88\small$\pm$0.17 & 73.24\small$\pm$0.20 & 85.60\small$\pm$0.64 & 81.85\small$\pm$0.07 & 90.61\small$\pm$0.26 & 92.75\small$\pm$0.11 & 86.41\small$\pm$0.76 \\

            \cmidrule{3-11}
            & & LaRa Det Raw & 96.69\small$\pm$0.12 & 83.49\small$\pm$0.12 & 73.69\small$\pm$0.32 & 86.46\small$\pm$0.15 & 81.30\small$\pm$0.10 & 91.12\small$\pm$0.27 & 93.19\small$\pm$0.14 & 86.56\small$\pm$0.51 \\

            \cmidrule{3-11}
            & & LaRa Prob Raw & 96.79\small$\pm$0.15 & 83.71\small$\pm$0.13 & 73.89\small$\pm$0.06 & 86.15\small$\pm$0.28 & 81.59\small$\pm$0.16 & 91.15\small$\pm$0.19 & 93.04\small$\pm$0.15 & 86.62\small$\pm$0.45 \\

            \cmidrule{3-11}
            & & LaRa MeDyate & 96.87\small$\pm$0.03 & 83.82\small$\pm$0.15 & 73.85\small$\pm$0.17 & 86.52\small$\pm$0.33 & 81.79\small$\pm$0.08 & 91.01\small$\pm$0.12 & 93.01\small$\pm$0.09 & 86.70\small$\pm$0.44 \\

            \cmidrule{2-11}
            & \multirow{6}{*}{633 859} & Prev MeDyate & 97.32\small$\pm$0.03 & 84.72\small$\pm$0.18 & 75.14\small$\pm$0.57 & 89.79\small$\pm$0.36 & 83.51\small$\pm$0.11 & 91.02\small$\pm$0.13 & 93.33\small$\pm$0.27 & 87.83\small$\pm$0.77 \\

            \cmidrule{3-11}
            & & LaRa TraDy & 97.31\small$\pm$0.04 & 84.90\small$\pm$0.08 & 74.36\small$\pm$0.27 & 86.56\small$\pm$0.89 & 83.63\small$\pm$0.12 & 91.08\small$\pm$0.17 & 92.72\small$\pm$0.13 & 87.22\small$\pm$0.97 \\

            \cmidrule{3-11}
            & & LaRa Det Raw & 97.19\small$\pm$0.03 & 84.55\small$\pm$0.13 & 74.25\small$\pm$0.15 & 87.67\small$\pm$0.54 & 83.20\small$\pm$0.03 & 91.13\small$\pm$0.27 & 92.62\small$\pm$0.28 & 87.23\small$\pm$0.70 \\

            \cmidrule{3-11}
            & & LaRa Prob Raw & 97.32\small$\pm$0.03 & 84.71\small$\pm$0.09 & 74.30\small$\pm$0.08 & 87.53\small$\pm$0.19 & 83.69\small$\pm$0.05 & 91.18\small$\pm$0.15 & 92.75\small$\pm$0.34 & 87.35\small$\pm$0.44 \\

            \cmidrule{3-11}
            & & LaRa MeDyate & 97.25\small$\pm$0.12 & 85.01\small$\pm$0.10 & 74.67\small$\pm$0.55 & 87.56\small$\pm$0.31 & 83.66\small$\pm$0.13 & 91.21\small$\pm$0.06 & 92.72\small$\pm$0.08 & 87.44\small$\pm$0.67 \\

            \cmidrule{2-11}
            & \multirow{6}{*}{2 767 686} & Prev MeDyate & 97.70\small$\pm$0.08 & 85.91\small$\pm$0.13 & 76.13\small$\pm$0.29 & 90.73\small$\pm$0.34 & 84.87\small$\pm$0.08 & 91.38\small$\pm$0.33 & 93.77\small$\pm$0.12 & 88.64\small$\pm$0.59 \\

            \cmidrule{3-11}
            & & LaRa TraDy & 97.52\small$\pm$0.07 & 85.81\small$\pm$0.12 & 75.22\small$\pm$0.06 & 88.21\small$\pm$0.46 & 84.70\small$\pm$0.02 & 91.23\small$\pm$0.18 & 93.33\small$\pm$0.07 & 88.00\small$\pm$0.52 \\

            \cmidrule{3-11}
            & & LaRa Det Raw & 97.54\small$\pm$0.02 & 85.69\small$\pm$0.21 & 75.42\small$\pm$0.57 & 90.26\small$\pm$0.22 & 84.83\small$\pm$0.06 & 91.04\small$\pm$0.37 & 93.52\small$\pm$0.22 & 88.33\small$\pm$0.78 \\

            \cmidrule{3-11}
            & & LaRa Prob Raw & 97.71\small$\pm$0.05 & 85.79\small$\pm$0.30 & 75.78\small$\pm$0.58 & 89.48\small$\pm$0.41 & 84.88\small$\pm$0.10 & 91.12\small$\pm$0.65 & 93.57\small$\pm$0.05 & 88.33\small$\pm$1.02 \\

            \cmidrule{3-11}
            & & LaRa MeDyate & 97.67\small$\pm$0.13 & 85.82\small$\pm$0.18 & 75.71\small$\pm$0.10 & 89.24\small$\pm$0.46 & 84.84\small$\pm$0.12 & 91.18\small$\pm$0.22 & 79.9\small$\pm$23.43 & 86.34\small$\pm$23.44 \\

            \cmidrule{2-11}
            & 31 889 952 & Baseline & 97.78\small$\pm$0.16 & 86.30\small$\pm$0.05 & 74.89\small$\pm$0.20 & 90.57\small$\pm$0.43 & 86.07\small$\pm$0.23 & 90.18\small$\pm$0.60 & 93.72\small$\pm$0.10 & 88.50\small$\pm$0.31 \\
            
            \bottomrule
        \end{tabular}
    }
\end{table*}
\begin{table*}
    \caption{Comparison of final top1 test accuracies between channel selection strategies with pretrained BERT and RoBERTa models, fine-tuned on various datasets and budgets.
    }
    \centering
    \renewcommand*{\arraystretch}{0.5}
    \resizebox{\textwidth}{!}{
    \label{tab:MeDyate_NLP}
        \begin{tabular}{ccccccc}
            \toprule
            \textbf{Model} & $B_{\text{mem}}$ & \textbf{Method} & \textbf{QNLI} & \textbf{RTE} & \textbf{SST2} & \textbf{Average} \\
            
            \midrule
            \multirow{48}{*}{BERT} 
            & \multirow{11}{*}{27 946} & Prev MeDyate & 84.38\small$\pm$0.06 & 58.24\small$\pm$1.82 & 89.37\small$\pm$0.13 & 77.33\small$\pm$1.83 \\

            \cmidrule{3-7}
            & & LaRa TraDy & 84.92\small$\pm$0.05 & 57.76\small$\pm$2.60 & 89.53\small$\pm$0.13 & 77.40\small$\pm$2.60 \\

            \cmidrule{3-7}
            & & LaRa Det Raw & 84.75\small$\pm$0.13 & 56.92\small$\pm$0.75 & 88.91\small$\pm$0.35 & 76.86\small$\pm$0.84 \\

            \cmidrule{3-7}
            & & LaRa Prob Raw & 84.62\small$\pm$0.30 & 57.04\small$\pm$1.30 & 88.91\small$\pm$0.18 & 76.86\small$\pm$1.35 \\

            \cmidrule{3-7}
            & & LaRa MeDyate & 84.58\small$\pm$0.07 & 58.48\small$\pm$1.65 & 89.18\small$\pm$0.24 & 77.41\small$\pm$1.67 \\

            \cmidrule{2-7}
            & \multirow{11}{*}{112 640} & Prev MeDyate &  84.51\small$\pm$0.25 & 58.48\small$\pm$0.63 & 89.53\small$\pm$0.33 & 77.51\small$\pm$0.75 \\

            \cmidrule{3-7}
            & & LaRa TraDy & 85.16\small$\pm$0.19 & 59.09\small$\pm$2.18 & 89.68\small$\pm$0.11 & 77.98\small$\pm$2.19 \\

            \cmidrule{3-7}
            & & LaRa Det Raw & 85.69\small$\pm$0.04 & 58.24\small$\pm$1.16 & 89.11\small$\pm$0.46 & 77.68\small$\pm$1.25 \\

            \cmidrule{3-7}
            & & LaRa Prob Raw & 85.28\small$\pm$0.11 & 57.52\small$\pm$0.55 & 89.56\small$\pm$0.00 & 77.45\small$\pm$0.56 \\

            \cmidrule{3-7}
            & & LaRa MeDyate & 85.19\small$\pm$0.21 & 56.56\small$\pm$0.21 & 89.41\small$\pm$0.37 & 77.05\small$\pm$0.47 \\

            \cmidrule{2-7}
            & \multirow{11}{*}{1 912 629} & Prev MeDyate & 86.78\small$\pm$0.43 & 55.72\small$\pm$1.63 & 90.37\small$\pm$0.34 & 77.62\small$\pm$1.72 \\

            \cmidrule{3-7}
            & & LaRa TraDy & 87.43\small$\pm$0.30 & 56.68\small$\pm$0.36 & 90.56\small$\pm$0.48 & 78.22\small$\pm$0.67 \\

            \cmidrule{3-7}
            & & LaRa Det Raw & 88.36\small$\pm$0.16 & 58.12\small$\pm$1.65 & 90.83\small$\pm$0.23 & 79.10\small$\pm$1.67 \\

            \cmidrule{3-7}
            & & LaRa Prob Raw & 88.31\small$\pm$0.10 & 58.12\small$\pm$2.25 & 90.63\small$\pm$0.66 & 79.02\small$\pm$2.35 \\

            \cmidrule{3-7}
            & & LaRa MeDyate & 87.79\small$\pm$0.10 & 55.84\small$\pm$1.04 & 90.44\small$\pm$0.76 & 78.02\small$\pm$1.29 \\

            \cmidrule{2-7}
            & \multirow{11}{*}{8 351 308} & Prev MeDyate & 89.05\small$\pm$0.14 & 61.13\small$\pm$1.99 & 91.21\small$\pm$0.66 & 80.46\small$\pm$2.10 \\

            \cmidrule{3-7}
            & & LaRa TraDy & 89.13\small$\pm$0.32 & 57.76\small$\pm$1.88 & 90.75\small$\pm$0.18 & 79.21\small$\pm$1.92 \\

            \cmidrule{3-7}
            & & LaRa Det Raw & 89.96\small$\pm$0.12 & 63.06\small$\pm$0.83 & 91.74\small$\pm$0.72 & 81.59\small$\pm$1.11 \\

            \cmidrule{3-7}
            & & LaRa Prob Raw & 89.69\small$\pm$0.14 & 61.01\small$\pm$0.96 & 91.48\small$\pm$0.13 & 80.73\small$\pm$0.98 \\

            \cmidrule{3-7}
            & & LaRa MeDyate & 89.95\small$\pm$0.18 & 61.49\small$\pm$1.71 & 90.90\small$\pm$0.53 & 80.78\small$\pm$1.80 \\

            \cmidrule{2-7}
            & 96 225 792 & Baseline & 90.81\small$\pm$0.27 & 62.45\small$\pm$1.81 & 91.74\small$\pm$0.50 & 81.67\small$\pm$1.90 \\

            \midrule
            \multirow{48}{*}{RoBERTa} 
            & \multirow{11}{*}{27 946} & Prev MeDyate & 89.69\small$\pm$0.06 & 57.04\small$\pm$0.72 & 93.31\small$\pm$0.07 & 80.01\small$\pm$0.73 \\

            \cmidrule{3-7}
            & & LaRa TraDy & 89.57\small$\pm$0.29 & 59.33\small$\pm$1.82 & 93.35\small$\pm$0.11 & 80.75\small$\pm$1.85 \\

            \cmidrule{3-7}
            & & LaRa Det Raw & 89.02\small$\pm$1.15 & 68.23\small$\pm$0.36 & 93.00\small$\pm$0.30 & 83.42\small$\pm$1.24 \\

            \cmidrule{3-7}
            & & LaRa Prob Raw & 89.42\small$\pm$0.22 & 66.91\small$\pm$3.47 & 93.08\small$\pm$0.54 & 83.14\small$\pm$3.52 \\

            \cmidrule{3-7}
            & & LaRa MeDyate & 89.81\small$\pm$0.27 & 63.18\small$\pm$4.72 & 93.31\small$\pm$0.18 & 82.10\small$\pm$4.73 \\

            \cmidrule{2-7}
            & \multirow{11}{*}{112 640} & Prev MeDyate & 90.05\small$\pm$0.09 & 60.41\small$\pm$2.29 & 93.39\small$\pm$0.18 & 81.28\small$\pm$2.30 \\

            \cmidrule{3-7}
            & & LaRa TraDy & 89.65\small$\pm$0.53 & 62.33\small$\pm$2.05 & 93.46\small$\pm$0.40 & 81.81\small$\pm$2.15 \\

            \cmidrule{3-7}
            & & LaRa Det Raw & 89.62\small$\pm$0.10 & 66.55\small$\pm$2.21 & 92.85\small$\pm$0.07 & 83.01\small$\pm$2.21 \\

            \cmidrule{3-7}
            & & LaRa Prob Raw & 89.66\small$\pm$0.15 & 66.55\small$\pm$2.35 & 92.89\small$\pm$0.34 & 83.03\small$\pm$2.38 \\

            \cmidrule{3-7}
            & & LaRa MeDyate & 89.77\small$\pm$0.33 & 67.75\small$\pm$3.28 & 93.27\small$\pm$0.13 & 83.60\small$\pm$3.30 \\

            \cmidrule{2-7}
            & \multirow{11}{*}{1 912 629} & Prev MeDyate & 91.40\small$\pm$0.06 & 75.45\small$\pm$0.72 & 93.92\small$\pm$0.53 & 86.92\small$\pm$0.90 \\

            \cmidrule{3-7}
            & & LaRa TraDy & 90.82\small$\pm$0.25 & 69.68\small$\pm$0.96 & 93.16\small$\pm$0.33 & 84.55\small$\pm$1.05 \\

            \cmidrule{3-7}
            & & LaRa Det Raw & 90.85\small$\pm$0.11 & 73.04\small$\pm$5.01 & 92.09\small$\pm$0.11 & 85.33\small$\pm$5.01 \\

            \cmidrule{3-7}
            & & LaRa Prob Raw & 90.87\small$\pm$0.19 & 75.21\small$\pm$1.46 & 92.39\small$\pm$0.46 & 86.16\small$\pm$1.54 \\

            \cmidrule{3-7}
            & & LaRa MeDyate & 90.60\small$\pm$0.47 & 74.61\small$\pm$1.85 & 92.97\small$\pm$0.07 & 86.06\small$\pm$1.91 \\

            \cmidrule{2-7}
            & \multirow{11}{*}{8 351 308} & Prev MeDyate & 91.52\small$\pm$0.41 & 75.21\small$\pm$1.63 & 93.16\small$\pm$0.46 & 86.63\small$\pm$1.74 \\

            \cmidrule{3-7}
            & & LaRa TraDy & 90.87\small$\pm$0.21 & 74.13\small$\pm$2.92 & 93.23\small$\pm$0.40 & 86.08\small$\pm$2.95 \\

            \cmidrule{3-7}
            & & LaRa Det Raw & 91.21\small$\pm$0.25 & 74.73\small$\pm$1.08 & 92.66\small$\pm$0.61 & 86.20\small$\pm$1.27 \\

            \cmidrule{3-7}
            & & LaRa Prob Raw & 90.67\small$\pm$0.24 & 74.37\small$\pm$0.72 & 92.93\small$\pm$0.07 & 85.99\small$\pm$0.76 \\

            \cmidrule{3-7}
            & & LaRa MeDyate & 90.49\small$\pm$0.78 & 70.52\small$\pm$3.51 & 92.97\small$\pm$0.40 & 84.66\small$\pm$3.62 \\

            \cmidrule{2-7}
            & 96 225 792 & Baseline & 92.31\small$\pm$0.14 & 76.41\small$\pm$0.55 & 93.16\small$\pm$0.92 & 87.29\small$\pm$1.08 \\
            
            \bottomrule            
        \end{tabular}
    }
\end{table*}

\end{document}